\documentclass[10pt,twocolumn,letterpaper]{article}

\usepackage{cvpr}              

\usepackage[pagebackref,breaklinks,colorlinks,allcolors=cvprblue]{hyperref}
\usepackage{multirow,diagbox}
\usepackage{graphicx}      
\usepackage{adjustbox}    
\usepackage[ruled,vlined,linesnumbered]{algorithm2e}
\usepackage{verbatim}
\usepackage{array,multicol}
\usepackage{float}
\usepackage{amsmath}
\usepackage{mathrsfs}
\usepackage{soul}
\usepackage{amssymb}
\usepackage{bbold}
\usepackage{tabularray}

\newcommand{\ours}{\textsc{FeatureFool}\xspace}

\definecolor{cvprblue}{rgb}{0.21,0.49,0.74}

\def\paperID{*****} 
\def\confName{CVPR}
\def\confYear{2026}

\title{FeatureFool: Zero-Query Fooling of Video Models via Feature Map}

\author{Duoxun Tang$^{1}$\thanks{Duoxun Tang is with Shenzhen International Graduate School, Tsinghua University, China. Email: tdx25@mails.tsinghua.edu.cn.} \hspace{0.1cm} Xi Xiao$^{1}$\thanks{Xi Xiao is the corresponding author at Shenzhen International Graduate School, Tsinghua University, China. Email: xiaox@sz.tsinghua.edu.cn. Guangwu Hu is with Shenzhen University of Information Technology, Shenzhen, China (e-mail: hugw@sziit.edu.cn).
Kangkang Sun is with the School of Astronautics, Harbin Institute of Technology, Harbin 150080, China (e-mail: kksun@hit.edu.cn).
Xiao Yang is with The Hong Kong University of Science and Technology (Guangzhou), Guangzhou, China (e-mail: xyang856@connect.hkust-gz.edu.cn).
Dongyang Chen is with Shenzhen International Graduate School, Tsinghua University, Shenzhen, China (e-mail: chen-dy25@mail.tsinghua.edu.cn).
Qing Li is with Peng Cheng Laboratory, Shenzhen, Guangdong 518038, China (e-mail: liq@pcl.ac.cn).
Yongjie Yin is with China Electronics Corporation, Beijing, China (e-mail: wingkit\_yyj@163.com).
Jiyao Wang is with The Hong Kong University of Science and Technology (Guangzhou), Guangzhou, China (e-mail: jwanggo@connect.ust.hk).} \hspace{0.1cm} Guangwu Hu$^{2}$ \hspace{0.1cm} Kangkang Sun$^{3}$ \hspace{0.1cm} Xiao Yang$^{6}$ \\
Dongyang Chen$^{1}$ \hspace{0.1cm} Qing Li$^{4}$ \hspace{0.1cm} Yongjie Yin$^{5}$ \hspace{0.1cm}  Jiyao Wang$^{6}$ \\\\
$^{1}$Tsinghua University \hspace{0.1cm} \\ 
$^{2}$Shenzhen University of Information Technology  ~~$^{3}$Harbin Institute of Technology, Shenzhen \hspace{0.1cm} \\ 
$^{4}$Peng Cheng Laboratory  ~~$^{5}$China Electronics Corporation \hspace{0.1cm} \\
$^{6}$Hong Kong University of Science and Technology, Guangzhou \\
}

\begin{document}
\maketitle
\begin{abstract} 
The vulnerability of deep neural networks (DNNs) has been preliminarily verified. Existing black-box adversarial attacks usually require multi-round interaction with the model and consume numerous queries, which is impractical in the real-world and hard to scale to recently emerged Video-LLMs. Moreover, no attack in the video domain directly leverages feature maps to shift the clean-video feature space. We therefore propose \ours, a stealthy, video-domain, \textbf{zero-query} black-box attack that utilizes information extracted from a DNN to alter the feature space of clean videos. Unlike query-based methods that rely on iterative interaction, \ours performs a zero-query attack by directly exploiting DNN-extracted information. This efficient approach is unprecedented in the video domain. Experiments show that \ours achieves an attack success rate above 70\% against traditional video classifiers without any queries. Benefiting from the transferability of the feature map, it can also craft harmful content and bypass Video-LLM recognition. Additionally, adversarial videos generated by \ours exhibit high quality in terms of SSIM, PSNR, and Temporal-Inconsistency, making the attack barely perceptible. \color{red}{This paper may contain violent or explicit content.}
\end{abstract}

\section{Introduction}
The rapid development of deep neural networks (DNNs) has achieved remarkable performance across numerous domains, yet adversarial attacks that craft imperceptible inputs can easily cause these models to behave abnormally \cite{goodfellow2014explaining, su2019one, qiu2019review, lin2020threats,liu2023adversarial, wan2024black, zneit2025adversarial, liu2025projattacker, zhou2025improving}.
Attacks on image-classification systems \cite{guo2019simple, jia2020adv, shi2021hyperspectral, croce2022sparse, tsai2023adversarial, ran2025black} mainly focused on iteratively perturbing single images. Videos, however, introduce an additional temporal dimension, requiring attackers to design frame-level perturbations; several methods thus covertly add adversaries to every frame \cite{yang2020patchattack, cao2024logostylefool, tang2024query}. The growth of computational power has fostered large models on which users increasingly rely, making their safety a critical concern. This trend has given rise to Large Vision–Language Models (LVLMs) \cite{li2022blip, team2023gemini, dai2023instructblip, chen2025janus} that combine visual encoders with large language models. Due to their multimodal nature, LVLMs expand the attack surface: adversaries can launch offensives from either the textual or the visual domain \cite{gao2024inducing, wang2025tapt, lyu2025pla, xie2025chain, medghalchi2025prompt2perturb}. For image-based LVLMs, for instance, PLA \cite{lyu2025pla} crafts adversarial prompts in the textual space to induce text-to-image generators to output pornographic content, while Verbose Image \cite{gao2024inducing} perturbs images to perform Denial-of-Service (DoS) attacks on LVLMs. The same vision–text fusion has further spawned Video-LLMs \cite{zhao2023learning, maaz2023video, weng2024longvlm, tang2025video, shu2025video} such as VideoLLaMA2 \cite{cheng2024videollama} and ShareGPT4Video \cite{chen2024sharegpt4video}, which outperform traditional architectures on video-understanding tasks \cite{li2024mvbench, zhou2025mlvu}. At the video-input level, adversaries typically craft perturbations on clean videos \cite{chen2022attacking, mu2024enhancing, cao2024logostylefool, tang2024query} to mislead conventional video classifiers \cite{tran2015learning, carreira2017quo} or simply replace or edit specific frames to assault Video-LLMs \cite{cao2025failures}. These approaches fall into white-box and black-box categories; the latter is more realistic and thus the focus of this paper.

\begin{figure}
\centering
	\captionsetup{
		font={scriptsize}, 
	}
	\begin{adjustbox}{valign=t}
		\includegraphics[width=1\linewidth]{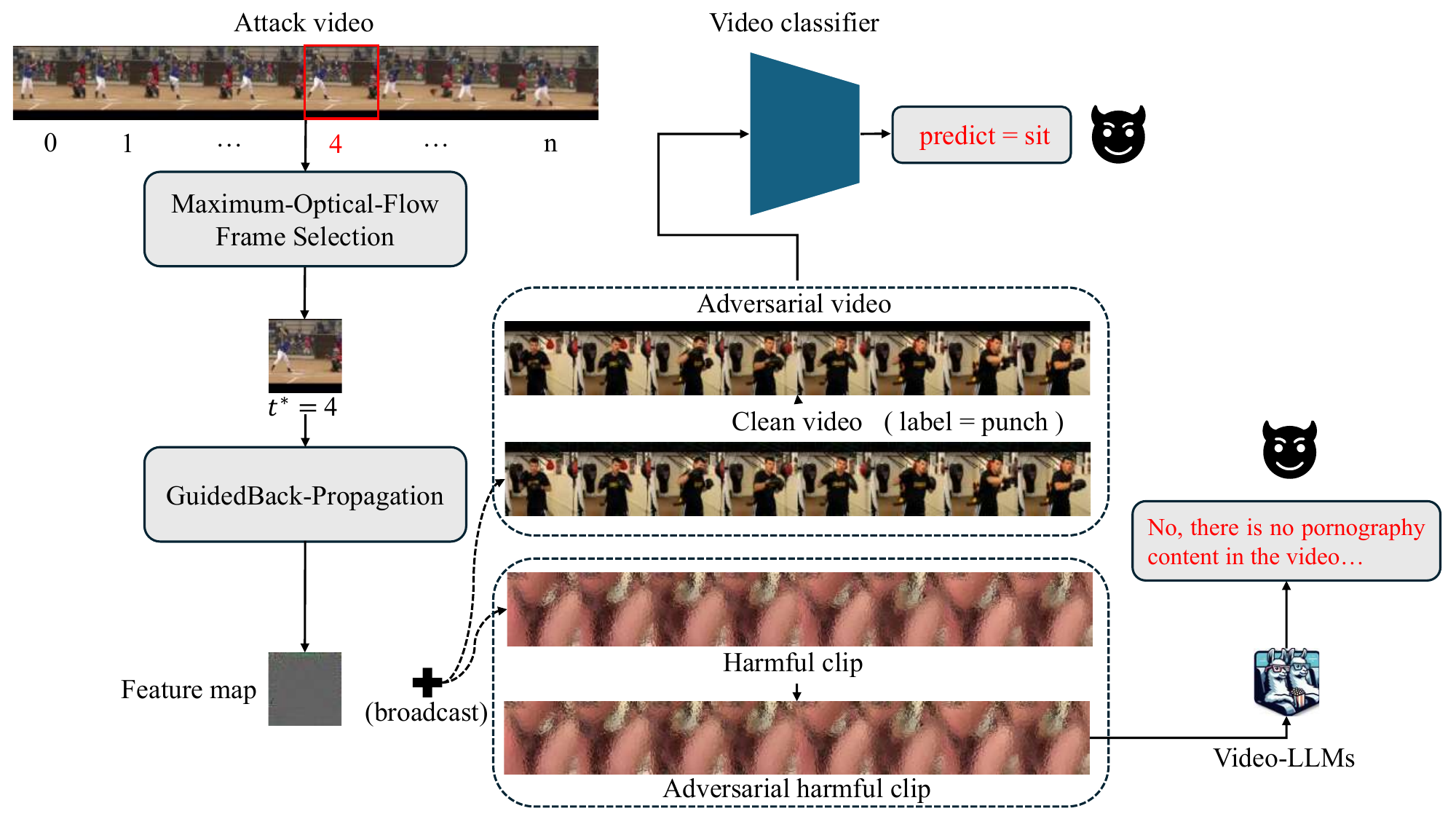}
	\end{adjustbox}
	\caption{The overview of FeatureFool: A \textbf{zero-query} video adversarial attack using only the feature map.}
	\label{figs:pipline}
    \vspace{-3mm}
\end{figure}

In black-box attacks against traditional DNNs classifiers, iterative methods \cite{li2021adversarial, wang2021reinforcement, pony2021over, wang2023global,rezghi2025tenad, song2025rlvs} repeatedly query the model, incurring heavy time and query costs. For example, Adv-watermark \cite{jia2020adv} relies on time-consuming heuristic search, while PatchAttack \cite{yang2020patchattack} employs reinforcement learning and often requires thousands of queries to succeed, an overhead unacceptable when scaling to recently emerged Video-LLMs. And these iterative methods may require \textbf{several hours} to complete a single attack on Video-LLMs. Although ZQBA \cite{costa2025zqba} has achieved zero-query attacks in the image domain, no works exploit feature-level perturbations for videos, and zero-query assaults remain under-explored for the video modality. A very recent method \cite{cao2025failures} proposes query-free tricks such as Frame Replacement Attack against Video-LLMs, yet these manipulations are visually conspicuous and can be easily filtered by human inspection.

To bridge the above gaps, we present \ours, a stealthy, \textbf{zero-query}, black-box attack that operates in the video domain and leverages information extracted from DNNs. 
Specifically, an attacker video is first processed by Maximum-Optical-Flow to locate the frame that carries the most critical motion information. Guided Back-propagation (GB) \cite{mostafa2022leveraging} is then applied to this frame to obtain a semantic and strong feature-map perturbation, which is broadcast to every frame of the victim video, neutralising the influence of differing frame-sampling strategies across Video-LLMs. Without any queries, \ours achieves an attack success rate (ASR) above 70\% against the C3D \cite{tran2015learning} and I3D \cite{carreira2017quo} video classifiers on HMDB-51 \cite{kuehne2011hmdb}, UCF-101 \cite{soomro2012ucf101} and Kinetics-400 \cite{kay2017kinetics}, while the adversarial samples exhibit high quality (SSIM \textgreater{} 0.87, PSNR \textgreater{} 28.00 dB). Moreover, harmful-content videos \cite{sultani2018real, elesawy2019real} crafted by \ours bypass the discrimination of Video-LLMs with a probability exceeding 70\% and can even induce hallucination. And remains robust against two state-of-the-art video-specific defenses (DPs \cite{lee2023defending} and TS \cite{hwang2024temporal}). An illustration of \ours is depicted in Figure~\ref{figs:pipline}. 

\textbf{Our contributions are:}
\begin{enumerate}
  \item We design an efficient \textbf{zero-query} black-box attack tailored for the video domain. To the best of our knowledge, this is the first work that leverages a feature map to attack-video systems without queries.
  \item We propose a novel pipeline that couples Maximum-Optical-Flow with Guided Back-propagation to extract the most influential feature map for perturbation.
  \item Extensive experiments on three benign video datasets, multi-category harmful clips, two mainstream video classifiers, and two powerful Video-LLMs demonstrate the vulnerability of both traditional classifiers and Video-LLMs to \ours.
\end{enumerate}

\section{Related Work}
\subsection{Black-box Adversarial Attack}
Early studies have shown that DNNs can be fooled by modifying only a single pixel \cite{su2019one}.
To better reflect real-world threats, numerous query-based black-box attacks have been proposed \cite{ilyas2018black, cheng2018query, guo2019simple, andriushchenko2020square}: Simba \cite{guo2019simple} achieves random perturbations in twenty lines of code; Andrew Ilyas et al. \cite{ilyas2018black} adopt Natural Evolution Strategies (NES) under query-limited or label-only settings; Square-Attack \cite{andriushchenko2020square} performs random search for efficient adversaries. For stealth, Adv-watermark \cite{jia2020adv} overlays a transparent, semantic watermark on images. These image-domain ideas inspire video attacks \cite{cao2023stylefool, cao2024logostylefool, li2024fmm, haghjooei2024qebb, mo2025query, yao2025stealthy}. Specifically, some video-oriented attacks pursue stealth by imposing stylised perturbations \cite{cao2023stylefool}, patch-edge constraints \cite{yang2020patchattack, cao2024logostylefool}, or temporally-sparse \cite{wei2019sparse} patterns on frames. These approaches, however, demand heavy prerequisites: intricate heuristics \cite{jia2020adv}, hundreds or even thousands of queries \cite{yang2020patchattack, cao2024logostylefool}, or even need to train a surrogate model \cite{wei2019sparse}. and temporally-sparse perturbations can be undermined by Video-LLMs’ varied sampling \cite{chen2024sharegpt4video,tang2025adaptive}, e.g. key-frame selection. ZQBA \cite{costa2025zqba} achieves training- and iteration-free attack in the image domain, yet its extension to video remains under-explored; how to better use frame information with feature maps is still an open issue, and its interaction with LLMs has not been investigated.

\subsection{Video Understanding Models}
Modern video understanding pipelines overwhelmingly rely on DNNs that stack spatio-temporal convolutions or attention layers to distill motion-aware embeddings \cite{tran2015learning, carreira2017quo, feichtenhofer2019slowfast, jiang2019stm, feichtenhofer2020x3d}. A classifier of this family ingests a clip, encodes long-range dynamics, and emits a categorical distribution over action labels. Representative backbones include SlowFast \cite{feichtenhofer2019slowfast}, which decouples low-speed spatial and high-speed temporal pathways; and X3D \cite{feichtenhofer2020x3d}, a channel-expansion recipe which inflates width and depth instead of resolution. Owing to their balanced accuracy-efficiency trade-off and open-source availability, C3D \cite{tran2015learning} and I3D \cite{carreira2017quo} remain the de-facto baselines for robustness evaluation; therefore we adopt them as victim models in our study.

Recently, Video-LLMs have emerged as a new paradigm that couples frozen visual encoders with LLMs, enabling open-ended text–video conversations, reasoning and safety filtering in a single unified architecture. Representative systems such as VideoLLaMA \cite{cheng2024videollama}, ShareGPT4Video \cite{chen2024sharegpt4video} and LLaVA-Video \cite{zhang2024video} have quickly pushed state-of-the-art results on video captioning, temporal grounding and visual question answering, while their multimodal alignment is typically achieved through lightweight adapters and large-scale image–video instruction tuning. For the forward process of Video-LLMs,
let $\mathcal F_\Theta$ denote a video-based large language model with parameters $\Theta=\{\phi,\mu\}$, composed of a visual encoder $f_\phi$ and a large language model $g_\mu$.  Given a video clip $X\in\mathbb{R}^{T\times C\times H\times W}$ and a user text query $Q_{\text{text}}$, the model proceeds as follows.  
To meet computational constraints, it first uniformly subsamples a frame set
\begin{equation}
\mathcal{V}_{s}=\{f_{t_{1}},f_{t_{2}},\dots,f_{t_{N}}\}\subset\mathcal{V},\quad N\ll T,
\end{equation}
from the full video $\mathcal{V}=\{f_{1},\dots,f_{T}\}$. A visual encoder $f_{\phi}$ then embeds these $N$ frames into tokens
\begin{equation}
\mathbf{Z}=f_{\phi}(\mathcal{V}_{s})\in\mathbb{R}^{L\times d},\quad
L=\frac{T}{\tau_{t}}\frac{H}{\tau_{h}}\frac{W}{\tau_{w}},
\end{equation}
where $L$ is the resulting number of spatio-temporal tokens. $\tau_{t},\tau_{h},\tau_{w}$ denote the temporal, height and width patch sizes, respectively, and $d$ is the output dimension of each visual token.
A template concatenates text query $Q_{\text{text}}$ with $\mathbf{Z}$ to form the prompt
\begin{equation}
\mathbf{S}=\bigl[\,\texttt{USER: }Q_{\text{text}}\;;\;\mathbf{Z}\;;\;\texttt{Assistant:}\bigr].
\end{equation}
Finally, the large language model $g_{\mu}$ generates the response token by token
\begin{equation}
y_{t}\sim p_{\Theta}(\,\cdot\,|\,\mathbf{S},y_{<t}),\quad t=1,\dots,K,
\end{equation}
yielding the complete answer $Y_{\text{respond}}=\{y_{t}\}_{t=1}^{K}$.
As these models are increasingly deployed for content moderation and interactive applications, understanding their robustness becomes imperative. This study operates purely in the visual domain: we perturb the clean video $X$ so that the Video-LLM $\mathcal F_\Theta$ fails to identify harmful content. Recent studies have revealed that adversarial inputs or content manipulations can cause Video-LLMs to bypass safety filters \cite{cao2025failures, cao2025poisoning}. Yet these methods are still easily detected by the human eye \cite{cao2025failures} or require queries and iteration \cite{cao2025poisoning}.

\section{Proposed Attack}
\subsection{Preliminary}
\textbf{Threat Model}. We consider a query-free black-box setting where the victim can be any public video classifier or Video-LLM (e.g., C3D \cite{tran2015learning}, I3D \cite{carreira2017quo}, VideoLLaMA2 \cite{cheng2024videollama}, ShareGPT4Video \cite{chen2024sharegpt4video}).  The adversary is allowed to extract feature maps from 3D-CNNs pretrained on open datasets (e.g., C3D, I3D) and to conduct attack offline on clean videos; no surrogate training, queries, or iterative feedback to the victim are permitted. 

\textbf{Attack Formulation.}
Let a video classifier
\begin{equation}
\phi:\mathbb{R}^{T\times C\times H\times W}\to\mathcal{Y}.
\end{equation}
An adversarial video $\mathbf{x}_{\mathrm{adv}}$ is generated by adding an imperceptible perturbation $\boldsymbol{\delta}$ to $\mathbf{x}$:
\begin{equation}
\mathbf{x}_{\mathrm{adv}}=\mathbf{x}+\boldsymbol{\delta},\qquad\|\boldsymbol{\delta}\|_{\infty}\le\varepsilon,
\end{equation}
where the budget $\varepsilon$ ensures human invisibility.
In practice, $\boldsymbol{\delta}$ is obtained by an $\ell_{\infty}$-ball projection,
so that the final adversarial video is
\begin{equation}
\mathbf{x}_{\mathrm{adv}}=\mathop{\mathrm{clip}}_{[0,1]}\!\bigl(\mathbf{x}+\boldsymbol{\delta}\bigr).
\end{equation}

The attack objective is to reduce the model's classification confidence, i.e., to achieve
\begin{equation}
\phi(\mathbf{x}_{\mathrm{adv}})\neq y.
\end{equation}

To maintain visual fidelity, the adversarial video must remain quantitatively indistinguishable from the original; {\em i.e.}, it should satisfy
\begin{equation}
\mathcal{D}(\mathbf{x},\mathbf{x}_{\mathrm{adv}})\in\mathcal{A},
\end{equation}
where $\mathcal{D}(\cdot,\cdot)$ denotes any appropriate perceptual metric ({\em e.g.}, SSIM, PSNR) and $\mathcal{A}$ is the acceptable value region for that metric.
Under this requirement, the perturbation $\boldsymbol{\delta}$ can be obtained in various ways. In \ours, we use feature map to fool video classifiers and evaluate the transferability of the resulting perturbations to Video-LLMs while maintaining \textbf{zero-query} and requiring no iterative optimization.

\subsection{Maximum Optical-Flow-based Grad Selection}
\label{sec:flow-grad}

\ours aims to inject a single, universal feature-map perturbation that pushes the clean video away from its original decision region in the 3D-CNN feature space.
To maximise the influence of this one-shot perturbation, we seek the frame that carries the most representative motion information.
Guided Back-propagation (GB) \cite{mostafa2022leveraging} introduces a masking mechanism during back-propagation that suppresses negative gradients, thereby producing a sharp visualisation of feature map which most strongly influence the model's decision.  It is therefore natural to use the high-impact feature map yielded by GB as the perturbation to be injected into the clean video. In the video domain, however, the introduction of the temporal dimension constitutes a fundamental departure from images \cite{costa2025zqba}: different frame-sampling strategies \cite{cheng2024videollama, maaz2023video} can directly undermine attack efficacy (e.g., the sampler may never draw the perturbed frames). Consequently, \ours adopts a global perturbations into every frame of the video.

We therefore propose to couple GB with a simple, zero-cost motion cue.
Empirically, frames with large flow magnitude often encapsulate more informative information \cite{li2024video}.
We thus identify the Maximum Optical-Flow frame and compute GB on that instant only. The selected gradient map is finally replicated along the temporal dimension to obtain a global perturbation. Formally, given an attack-video tensor $\mathbf{X}^{\mathrm{att}}\in\mathbb{R}^{C\times T\times H\times W}$, we first compute the dense optical flow between every pair of consecutive frames using the Farneback algorithm \cite{farneback2003two}:

\begin{equation}
\mathcal{F}_{t}(\mathbf{p})=\texttt{FB}(\mathbf{X}_{t-1}^{\mathrm{att}},\mathbf{X}_{t}^{\mathrm{att}})(\mathbf{p})\in\mathbb{R}^{2},\;
\mathbf{p}\in\Omega,\,t=2,\dots,T,
\end{equation}
where $\mathcal{F}_{t}(\mathbf{p})=[\Delta u,\Delta v]$ denotes the sub-pixel displacement from frame $t-1$ to frame $t$ at pixel $\mathbf{p}$, and $\Omega\subset\mathbb{R}^{2}$ denotes the image plane. The horizontal and vertical displacement fields are obtained by minimising the quadratic polynomial error
\begin{equation}
\Delta u_{t}(\mathbf{p})=\sum_{\mathbf{q}\in\mathcal{N}(\mathbf{p})}w(\mathbf{q})\bigl[I_{t}(\mathbf{q})-I_{t-1}(\mathbf{q})\bigr]\frac{\partial I_{t-1}}{\partial x}\!(\mathbf{q}),
\end{equation}

\begin{equation}
\Delta v_{t}(\mathbf{p})=\sum_{\mathbf{q}\in\mathcal{N}(\mathbf{q})}w(\mathbf{q})\bigl[I_{t}(\mathbf{q})-I_{t-1}(\mathbf{q})\bigr]\frac{\partial I_{t-1}}{\partial y}\!(\mathbf{q}),
\end{equation}
with $I_{t}(\mathbf{p})=\mathbf{X}_{t}^{\mathrm{att}}(\mathbf{p})$ the intensity at pixel $\mathbf{p}$, $\mathcal{N}(\mathbf{p})$ a $5\times 5$ neighbourhood centred at $\mathbf{p}$, and $w(\mathbf{q})$ Gaussian weights.

Each vector $\mathcal{F}_{t}(h,w)$ gives the displacement $(\Delta u,\Delta v)$ of pixel $(h,w)$.
The average magnitude of frame $t$ is

\begin{equation}
m_{t}=\frac{1}{HW}\sum_{h,w}\sqrt{(\Delta u_{h,w})^{2}+(\Delta v_{h,w})^{2}}.
\end{equation}
Then append $m_{0}\!=\!m_{1}$ and $m_{T}\!=\!m_{T-1}$ to handle boundaries, and select the index with
\begin{equation}
t^{*}=\arg\max_{t}\;m_{t}.
\end{equation}

\subsection{Feature Map Perturbation}

\begin{figure}
\centering
	\captionsetup{
		font={scriptsize}, 
	}
	\begin{adjustbox}{valign=t}
		\includegraphics[width=0.8\linewidth]{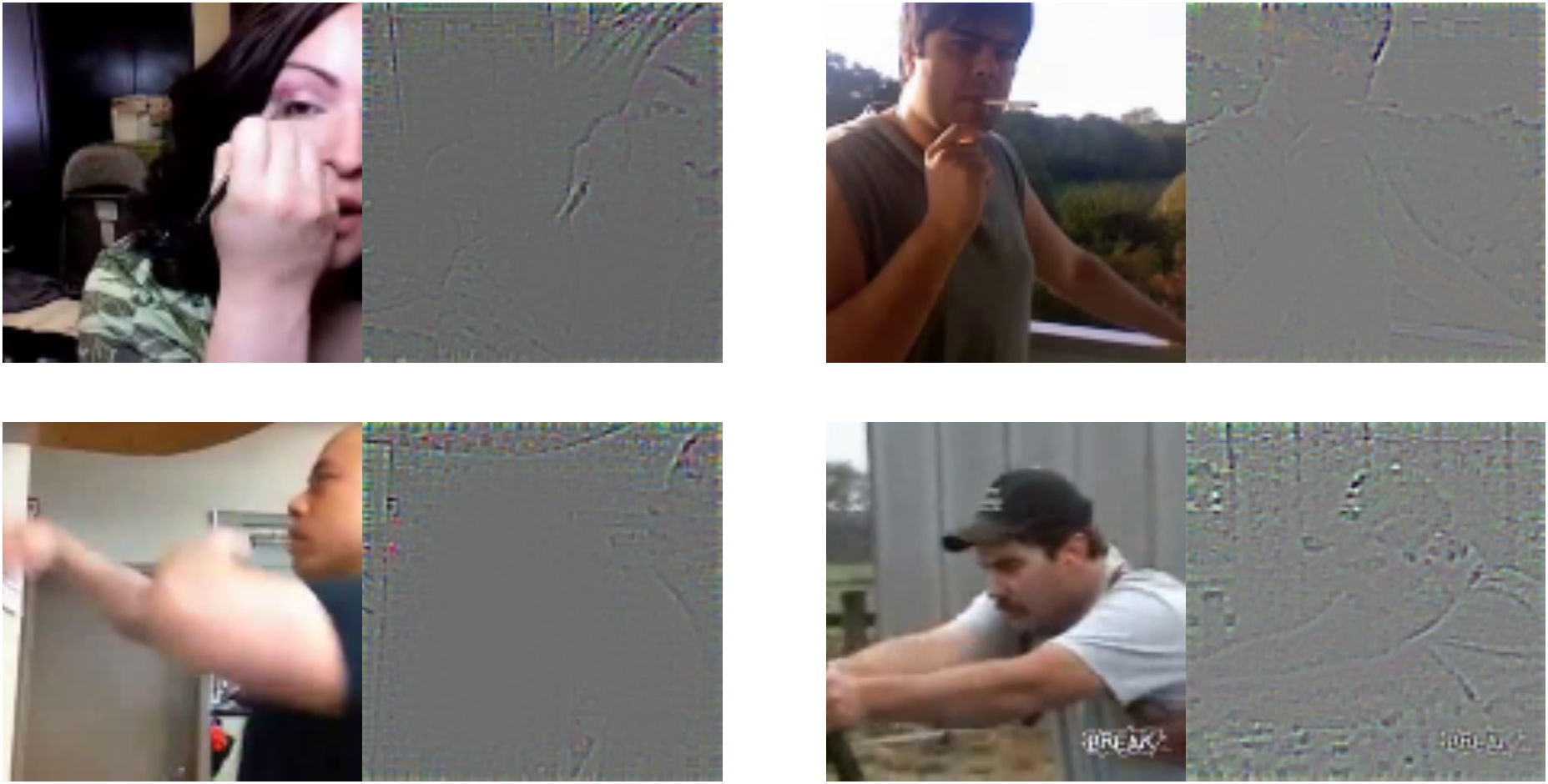}
	\end{adjustbox}
	\caption{Feature map extracted by Guided Back-propagation from the max-flow frame.}
	\label{figs:feature_map}
    \vspace{-3mm}
\end{figure}

GB is then applied only at frame $t^{*}$.
Let $\phi_{\ell}(\cdot;\theta)$ denote the classifier up to layer $\ell$; the gradient map used for perturbation is
\begin{equation}
\mathbf{G}=\nabla_{\mathbf{X}_{t^{*}}^{\mathrm{att}}}\,\phi_{\ell}(\mathbf{X};\theta)),
\end{equation}
where the backward pass is modified by Guided-ReLU \cite{springenberg2014striving} to suppress negative gradients.
Specifically, the standard ReLU backward mask
\begin{equation}
\mathbb{1}_{\!\text{ReLU}}=\mathbb{1}_{\frac{\partial\mathcal{L}}{\partial\mathbf{z}}>0}\cdot\mathbb{1}_{\mathbf{z}>0}
\end{equation}
is replaced by
\begin{equation}
\mathbb{1}_{\!\text{GReLU}}=\mathbb{1}_{\frac{\partial\mathcal{L}}{\partial\mathbf{z}}>0}\cdot\mathbb{1}_{\mathbf{z}>0}\cdot\mathbb{1}_{\text{grad}_{\text{in}}>0},
\end{equation}
ensuring that only positive gradients w.r.t.\ both the activation $\mathbf{z}$ and the incoming gradient $\text{grad}_{\text{in}}$ are back-propagated, yielding a sharper, discriminative gradient map $\mathbf{G}\in\mathbb{R}^{H\times W\times C}$. An illustrative example of the extracted feature map is shown in Figure~\ref{figs:feature_map}.
Finally, the universal perturbation is broadcast to all frames:
\begin{equation}
\mathbf{X}_{\mathrm{adv}}=\mathbf{X}+\alpha\,\mathbf{G}^{\!\rightarrow\!\mathrm{T}},\qquad\|\mathbf{X}_{\mathrm{adv}}-\mathbf{X}\|_{\infty}\le\varepsilon,
\end{equation}
with $\alpha$ chosen to satisfy the $\ell_{\infty}$ budget and $\mathbf{G}^{\!\rightarrow\!\mathrm{T}}$ denoting replication of $\mathbf{G}$ along the temporal axis.
This optical-flow-guided selection ensures that the single-frame gradient carries motion-rich, model-sensitive information, yielding a strong yet imperceptible adversarial video. The algorithmic overview of \ours is given in Algorithm~\ref{alg:featurefool}.

\begin{algorithm}[t]
   \caption{\ours: Zero-Query Video Adversarial Attack}
   \label{alg:featurefool}
   \KwIn{clean video $\mathbf{X}\in\mathbb{R}^{C\times T\times H\times W}$;
         pretrained 3D-CNN $\phi(\cdot;\theta)$;
         attack-video $\mathbf{X}^{\mathrm{att}}$ (any source);
         layer $\ell$;
         budget $\varepsilon$;
         scale $\alpha$.}
   \KwOut{adversarial video $\mathbf{X}_{\mathrm{adv}}\!\in\![0,1]^{C\times T\times H\times W}$.}
   compute optical-flow magnitudes $m_{t}$ for $\mathbf{X}^{\mathrm{att}}$ 
   $t^{*}=\arg\max_{t}\;m_{t}$\;
   replace ReLU by GReLU in $\phi(\cdot;\theta)$\;
   forward $\mathbf{X}^{\mathrm{att}}$ through $\phi(\cdot;\theta)$\;
   backward w.r.t.\ logits of corresponding class\;
   extract gradient map $\mathbf{G}\!\gets\!\nabla_{\mathbf{X}^{\mathrm{att}}_{t^{*}}}\phi_{\ell}(\mathbf{X}^{\mathrm{att}};\theta)$\;
   restore original ReLU\;
   $\mathbf{G}\gets\text{clip}_{[0,1]}(\text{ReLU}(\mathbf{G}))$\;
   $\boldsymbol{\delta}\gets\alpha\cdot\mathbf{G}$\;
   $\boldsymbol{\delta}\gets\text{clip}_{[-\varepsilon,\varepsilon]}(\boldsymbol{\delta})$\;
   replicate $\boldsymbol{\delta}$ along time: $\boldsymbol{\Delta}\!\gets\!\boldsymbol{\delta}^{\rightarrow T}$\;
   $\mathbf{X}_{\mathrm{adv}}\gets\text{clip}_{[0,1]}(\mathbf{X}+\boldsymbol{\Delta})$\;
   \Return{$\mathbf{X}_{\mathrm{adv}}$}
\end{algorithm}

\begin{figure}[ht]
\centering
	\captionsetup{
			font={scriptsize}, 
		}
	\centerline{\includegraphics[width=1.0\hsize]{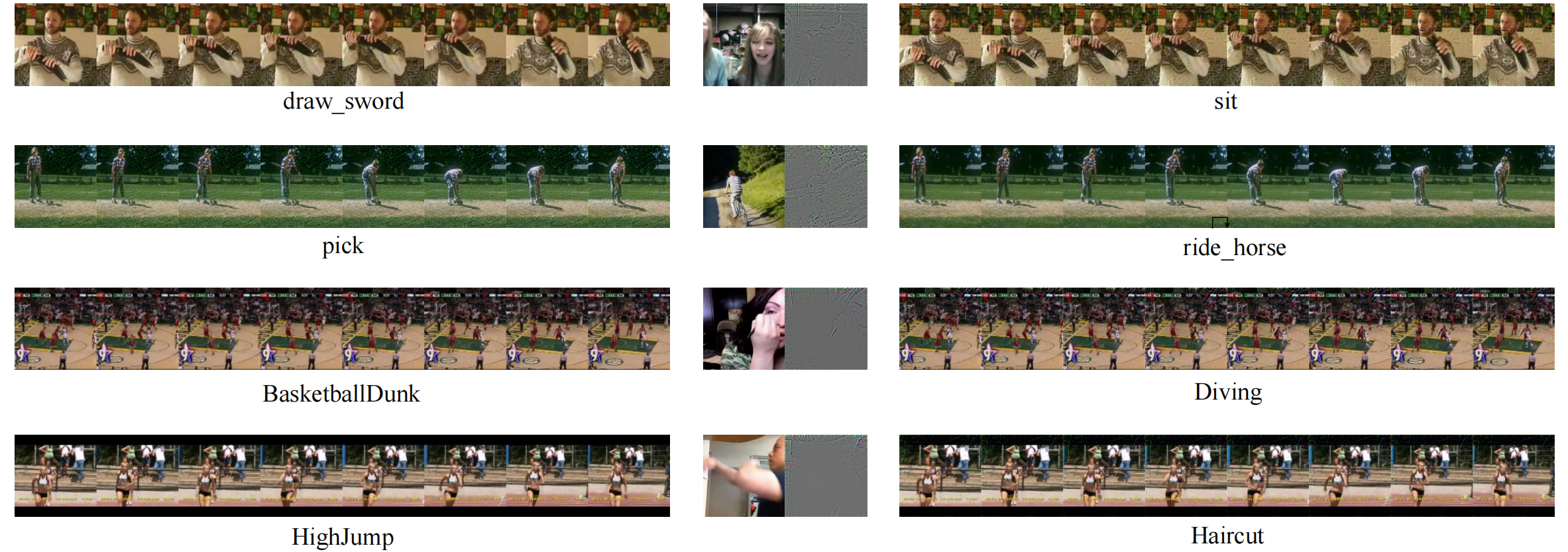}}
	\caption{Adversarial examples of \ours. Left: clean video; middle: attack medium; right: adversarial video.}
	\label{figs:Ours}
    \vspace{-3mm}
\end{figure}
\begin{table*}[ht]
\centering
\caption{Attack performance comparison on video classifiers.}
\label{tab:attack_performance_classifier}
\footnotesize
\resizebox{0.95\linewidth}{!}{
\begin{tabular}{cc rrrrr rrrrr rrrrr}
\toprule
\multirow{2}{*}{Model} & 
\multirow{2}{*}{Attack} & \multicolumn{5}{c}{UCF-101} &  \multicolumn{5}{c}{HMDB-51} & \multicolumn{5}{c}{Kinetics-400}\\
\cmidrule(lr){3-7} \cmidrule(lr){8-12} \cmidrule(l){13-17}
 & & ASR$\uparrow$ & \#Queries$\downarrow$ & TI$\downarrow$ & SSIM$\uparrow$ & PSNR$\uparrow$ & ASR$\uparrow$ & \#Queries$\downarrow$ & TI$\downarrow$ & SSIM$\uparrow$ & PSNR$\uparrow$ &
ASR$\uparrow$ & \#Queries$\downarrow$ & TI$\downarrow$ & SSIM$\uparrow$ & PSNR$\uparrow$\\
\midrule
\multirow{5}{*}{C3D} 
& Sparse-RS \cite{croce2022sparse} & 62\% & 1 & 5.6834 & 0.7677 & 4.3023 &  52\% & 1 & 6.7134 & 0.7036 & 5.3265 & 58\% & 1 & 8.3124 & 0.7518 & 5.6029\\
& Adv-watermark \cite{jia2020adv} & 58\% & 668.4  & 4.7701 & 0.8267 & 10.0632 & 49\% &  466.9 & 4.6522 & 0.8069 & 11.6475 & 56\% & 716.9 & 4.6202 & 0.8669 & 11.9548\\
& PatchAttack \cite{yang2020patchattack} & 68\% & 6,523.7  & 65.2708 & 0.7264 & 6.7248 & 65\% &  5,564.1 & 107.5264 & 0.7492 & 7.5482 & 52\% & 6,112.4 & 88.1057 & 0.8669 & 7.0154\\
& BSC \cite{chen2022attacking} & \textbf{72\%} & 3,968.4  & 4.2006 & 0.8519 & 15.7562 & 71\% &  3,645.9 & 3.9673 & 0.8618 & 16.6482 & \textbf{76\%} & 5,102.9 & 4.8526 & 0.8547 & 16.3519\\
& \textbf{FeatureFool} & 70\% & \textbf{0} & \textbf{3.1664} & \textbf{0.8834} & \textbf{29.0297} & \textbf{72\%} & \textbf{0} & \textbf{3.6821} & \textbf{0.8755} &  \textbf{28.9602} 
& 70\% & \textbf{0} & \textbf{3.7553} & \textbf{0.8864} &  \textbf{28.5624}\\ 
\midrule
\multirow{5}{*}{I3D} 
& Sparse-RS \cite{croce2022sparse} & 53\% & 1 &  5.7934 & 0.7351 & 4.6163 & 54\% &  1 & 6.6107 & 0.7662 & 4.9267 & 53\% & 1 & 7.3950 & 0.7629 & 6.0248\\
& Adv-watermark \cite{jia2020adv} & 61\% & 768.8 & 4.7364 & 0.8362 & 9.8469 & 57\% & 500.5 &  5.2634 & 0.8264 & 10.2657 & 55\% & 346.2 & 5.0264 & 0.8362 & 11.6592\\
& PatchAttack \cite{yang2020patchattack} & 71\% & 6,476.9 & 157.9572 & 0.7410 & 6.9507 & 69\% & 5,981.5 &  112.8429 & 0.7049 & 5.9014 & 62\% & 4,625.0 & 153.4861 & 0.7214 & 5.9871\\
& BSC \cite{chen2022attacking} & 70\% & 4,952.1 & 4.6657 & 0.8422 & 14.6237 & \textbf{74\%} & 3,017.2 &  4.9438 & 0.8322 & 14.9951 & 70\% & 4,722.6 & 4.1176 & 0.8493 & 17.9921\\
& \textbf{FeatureFool} & \textbf{74\%} & \textbf{0} & \textbf{3.2937} & \textbf{0.8914} & \textbf{30.0137} & 73\% & \textbf{0} & \textbf{4.3467} & \textbf{0.8861} & \textbf{29.4111} 
& \textbf{72\%} & \textbf{0} & \textbf{3.6594} & \textbf{0.8731} & \textbf{30.2497}\\
\bottomrule
\end{tabular}
} 
\end{table*}

\section{Experiments}
\subsection{Experimental Setup}
\textbf{Video Datasets.}
Datasets comprise 100 test videos each from UCF-101 \cite{soomro2012ucf101}, HMDB-51 \cite{kuehne2011hmdb} and Kinetics-400 \cite{kay2017kinetics}, all randomly sampled and verified to be correctly classified by the victim model.  
In addition, we collect 20 clips for each of the three harmful categories: violence, crime  and pornography, yielding 60 videos in total to evaluate attack transferability on Video-LLMs.
Violence and crime clips are taken from the Real-Life Violence Situations Dataset \cite{elesawy2019real} and the UCF-Crime Dataset \cite{sultani2018real}. These publicly available datasets are mainly sourced from online platforms including YouTube and Pornhub. More information in the Sup. Mat \ref{sec:Datasets}.

\textbf{Victim Models.}
We evaluate against two representative video classifiers, C3D \cite{tran2015learning} and I3D \cite{carreira2017quo}. C3D applies 3-D convolutions to encode spatio-temporal patterns directly, whereas I3D inflates 2-D ImageNet weights and exploits optical-flow streams to capture inter-frame dynamics; both yield strong classification accuracy.
Every input clip is resized to 16 frames of $112\times112$ pixels (Kinetics-400 in I3D is 32 frames). We further test VideoLLaMA2 \cite{cheng2024videollama} and ShareGPT4Video \cite{chen2024sharegpt4video}, which integrate large-scale language priors with visual encoders for open-ended video understanding. Please refer to the Sup. Mat \ref{sec:Victim Models}. for more information.

\textbf{Baselines.}  
We compare with four query-efficient attacks:  
\textbf{Adv-watermark} \cite{jia2020adv} overlays an optimized translucent watermark, \textbf{Sparse-RS} \cite{croce2022sparse} add patches based on random search, \textbf{PatchAttack} \cite{yang2020patchattack} adds RGB perturbations and \textbf{BSC} \cite{chen2022attacking} perturbs the video in the form of transparent bullet-screen comments. 

\textbf{Metrics}.
We evaluate attack effectiveness with five metrics.  
\textbf{Attack Success Rate (ASR)} measures the proportion of videos for which either (i) the action-classifier predicts an incorrect label or (ii) the Video-LLM fails to identify harmful content (violence, crime, pornography).  Owing to the potential overlap between violent and criminal acts, we employ a unified prompt---\textit{``Is there any violent or criminal content in the video?''}---to interrogate the Video-LLM. For pornographic clips, we adopt a similar prompt \textit{``Is there any pornographic content in the video?''}.
\textbf{\#Queries} counts the number of model inferences required to craft an adversarial video; lower is better for black-box efficiency.  
\textbf{Temporal Inconsistency (TI)} \cite{ruder2018artistic} quantifies the temporal flicker of adversarial videos.
Perceptual quality is quantified by \textbf{SSIM} \cite{wang2004image} and \textbf{PSNR} between the adversarial and the original video; higher values indicate smaller visual distortion.  
Arrows ($\uparrow/\downarrow$) denote desirable directions for each metric. More metric details in the Sup. Mat \ref{sec:Metrics}.

\begin{figure}[ht]
\centering
	\captionsetup{
			font={scriptsize}, 
		}
	\centerline{\includegraphics[width=1\hsize]{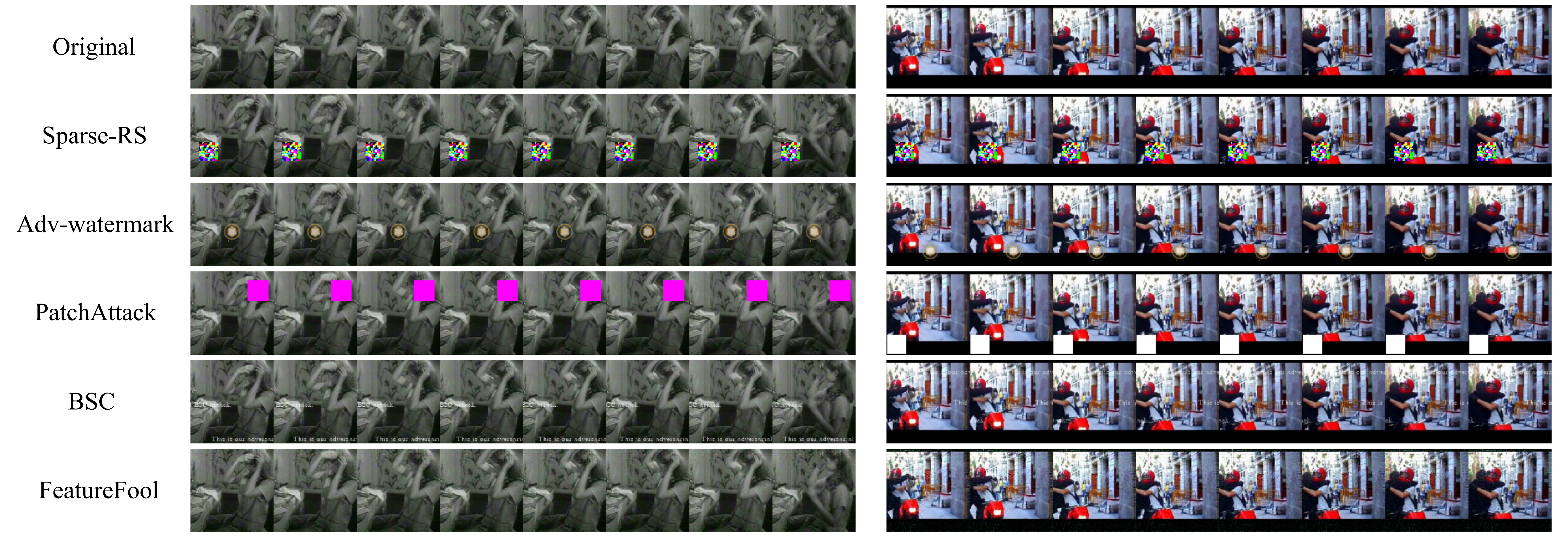}}
	\caption{Qualitative comparison of different methods.}
	\label{figs:Compare}
    \vspace{-3mm}
\end{figure}

\begin{figure}[ht]
\centering
	\captionsetup{
			font={scriptsize}, 
		}
	\begin{adjustbox}{valign=t}
	\includegraphics[width=0.95\linewidth]{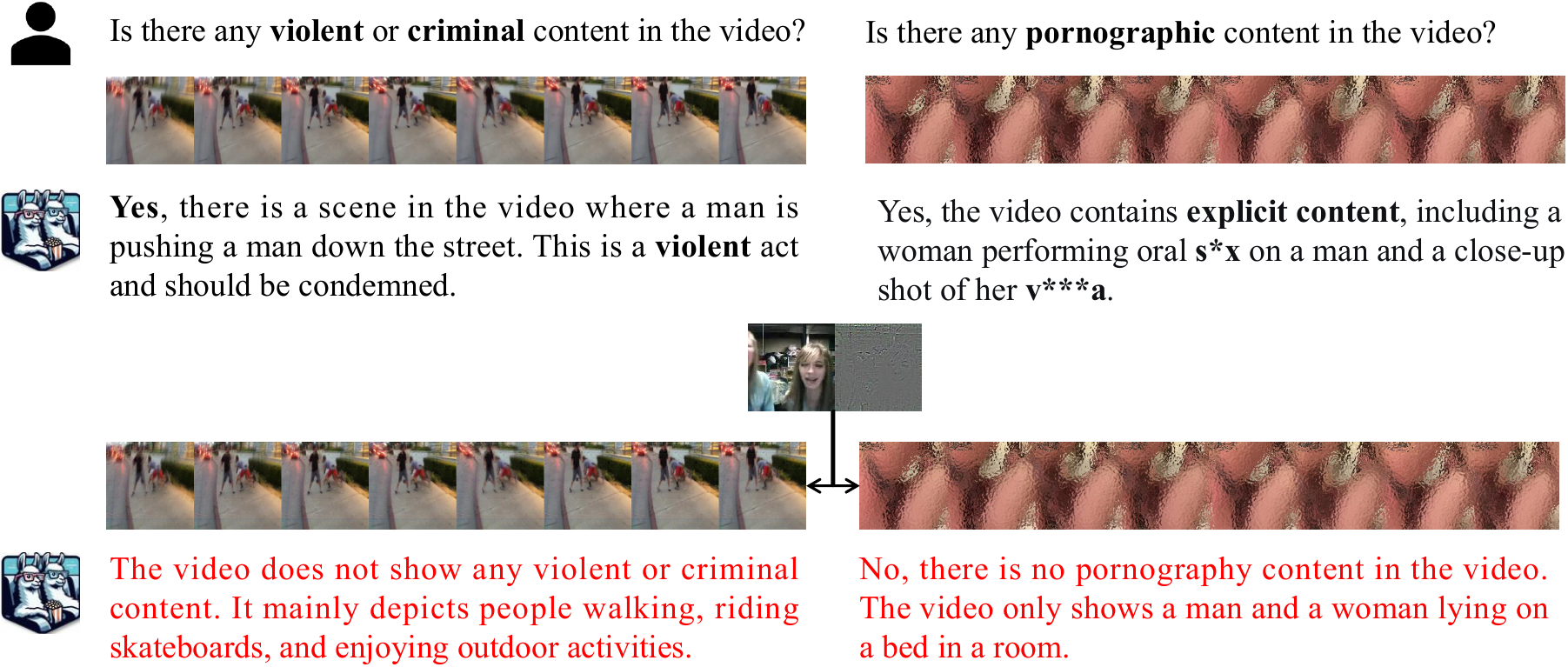}
	\end{adjustbox}
	\caption{Harmful content crafted by \ours can bypass the judgment of Video-LLM.}
	\label{figs:harmful}
    \vspace{-3mm}
\end{figure}

\subsection{FeatureFool Attack Performance}
\textbf{Performance on Video Classifiers.}
Figure~\ref{figs:Ours} illustrates \ours misleading the original video category into four distinct non-ground-truth classes under different attack media. Table~\ref{tab:attack_performance_classifier} summarises the cross-dataset and cross-model performance of the proposed \ours against two black-box, query-efficient baselines. Under such an extreme query budget, Sparse-RS fails to achieve a high success rate and, moreover, introduces the largest degradation in video quality, yielding the worst SSIM and PSNR values in all cases. This low-quality adversarial video is most likely caused by the semantically meaningless patches introduced by Sparse-RS. For Adv-watermark, its ASR is comparable to Sparse-RS, and the resulting adversarial videos are visually better, yet still far behind \ours. PatchAttack and BSC require a large number of queries, and the adversarial quality of PatchAttack is poor.
Specifically, \ours attains the highest ASR with \textbf{zero} queries and produces adversarial videos of the highest quality: its PSNR is 2--3$\times$ higher than that of Adv-watermark, while its SSIM consistently exceeds 0.87, indicating that the attack remains virtually imperceptible to the human eye even at the highest success rate. These benefit from \ours leverages the strong motion information carried by the maximum-flow frame and, via GB, appropriately highlights the feature map. Such a semantically meaningful, global perturbation (feature map) exerts a greater impact on model predictions without any queries. Figure~\ref{figs:Compare} provides a qualitative comparison of the three attacks, showing intuitively that \ours is the most stealthy and barely perceptible to the naked eye.

\begin{table}[!t]  
\centering
\caption{Attack Performance on Video-LLMs. Metric: ASR (\%).
}
\label{tab:attack_performance_llm}
\resizebox{0.95\linewidth}{!}{
\begin{tabular}{cccccc}
\toprule
\multirow{2}{*}{\textbf{Attack}} 
& \multirow{2}{*}{\textbf{Model}} 
& \multicolumn{2}{c}{\textbf{Harmful Category}} & \multirow{2}{*}{\textbf{Avg} $\uparrow$} 
\\
\cline{3-4}
& & \rule{0pt}{2.5ex}Violence \& Crime & Pornography \\
\midrule
\multirow{2}{*}{\shortstack{Sparse-RS \cite{croce2022sparse}}} 
& VideoLLaMA & 12.50\% & 10.00\%  & 11.25\% \\
& ShareGPT4Video & 10.00\% & 15.00\% & 12.50\% \\
\midrule
\multirow{2}{*}{\shortstack{Adv-watermark \cite{jia2020adv}}}
& VideoLLaMA & 7.50\% & 10.00\%  & 12.50\% \\
& ShareGPT4Video & 12.50\% & 20.00\% & 16.25\% \\
\midrule
\multirow{2}{*}{\shortstack{PatchAttack \cite{yang2020patchattack}}}
& VideoLLaMA & 24.5\% & 30.00\%  & 27.25\% \\
& ShareGPT4Video & 27.50\% & 44.00\% & 35.75\% \\
\midrule
\multirow{2}{*}{\shortstack{BSC \cite{chen2022attacking}}}
& VideoLLaMA & 30.00\% & 40.00\%  & 35.00\% \\
& ShareGPT4Video & 35.50\% & 33.00\% & 34.25\% \\
\midrule
\multirow{2}{*}{\shortstack{FeatureFool}}
& VideoLLaMA & \textbf{77.50\%} & \textbf{75.00\%}  & \textbf{76.25\%} \\
& ShareGPT4Video & \textbf{72.50\%} & \textbf{70.00\%} & \textbf{71.25\%} \\
\bottomrule
\end{tabular}
}
\end{table}

\textbf{Performance on Video-LLMs.}
Table~\ref{tab:attack_performance_llm} reports the ASR against two VideoLLaMA2 and ShareGPT4Video on 60 harmful clips. Due to the inherent overlap between violent and criminal content, the ASR for these two categories is computed jointly. We observe that \ours can effectively bypass the harmful-content detection capabilities of both models: for violence-, crime-, and pornography-related clips modified by \ours's feature map, more than 70\% of the samples are judged as "free of the corresponding harmful content." In contrast, Sparse-RS, Adv-watermark, PatchAttack and BSC are clearly ineffective against Video-LLMs, as they fail to exert sufficient influence on the victim video's feature space. This gap confirms that feature-rich perturbation are far more influential than unstructured or weakly semantic perturbations when attacking Video-LLMs. Figure~\ref{figs:harmful} demonstrates the behaviour of Video-LLM under a successful attack. \ours can also trigger hallucinations in Video-LLMs, an example is given in Sup. Mat \ref{sec:Hallucination}.

\textbf{Cross architectures evaluation.} Table~\ref{tab:transfer_architecture} reports the ASR of \ours under different model pairs. It can be observed that using a feature map extracted from a source model different from the victim model has a noticeable yet acceptable impact on attack performance, which is attributed to architectural discrepancies. These results fully demonstrate the good cross-model transferability of \ours.

\begin{table}[!t]
  \centering
  \caption{Cross-architecture transfer ASR (\%). \textbf{S} denotes the source model that generates the feature map, and \textbf{V} reprets the victim model under attack. All sources and victims share the same pretrained datasets (UCF-101, HMDB-51 and Kinetics-400).}
  \label{tab:transfer_architecture}
  \resizebox{0.95\linewidth}{!}{
  \begin{tabular}{lcccc}
    \toprule
    Dataset     & \multicolumn{2}{c}{C3D (S)} & \multicolumn{2}{c}{I3D (S)} \\
    \cmidrule(lr){2-3} \cmidrule(lr){4-5}
                & C3D ($V$) & I3D ($V$) & I3D ($V$) & C3D ($V$)\\
    \midrule
    UCF-101     & \textbf{70.0\%} & 66.0\% & \textbf{74.0\%} & 68.0\%  \\
    HMDB-51     & \textbf{72.0\%} & 64.0\% & \textbf{73.0\%} & 65.0\%  \\
    Kinetics-400& \textbf{70.0\%} & 65.0\% & \textbf{72.0\%} & 67.0\%  \\
    \bottomrule
  \end{tabular}
  }
\end{table}

\textbf{Cross datasets evaluation.} Table~\ref{tab:transfer_datasets} presents \ours’s performance across different pre-training datasets. When the victim model’s pre-trained dataset differs from that of the source model, the drop in ASR is smaller than the degradation caused by architectural mismatch. In all cases, the attack remains highly effective, making the \textbf{zero-query} performance highly satisfactory.

\begin{table}[t]
  \centering
  \caption{Cross-dataset transfer ASR (\%). \textbf{S} denotes the source model that generates the feature map, and \textbf{V} reprets the victim model under attack. All sources and victims share the same architecture (C3D or I3D).}
  \label{tab:transfer_datasets}
  \resizebox{\linewidth}{!}{
    \begin{tabular}{lcccc}
      \toprule
      \multirow{2}{*}{Source Model ($S$)} & \multicolumn{3}{c}{Victim Model ($V$)} \\
      \cmidrule(lr){2-4}
                                          & UCF-101 & HMDB-51 & Kinetics-400 \\
      \midrule
      C3D-UCF-101 & \textbf{70.0\%} & 68.0\% & 64.0\% \\
      C3D-HMDB-51 & 69.0\% & \textbf{72.0\%} & 66.0\% \\
      C3D-Kinetics-400 & 68.0\% & 66.0\% & \textbf{70.0\%} \\
      \midrule
      I3D-UCF-101 & \textbf{74.0\%} & 70.0\% & 67.0\% \\
      I3D-HMDB-51 & 69.0\% & \textbf{73.0\%} & 68.0\% \\
      I3D-Kinetics-400 & 66.0\% & 68.0\% & \textbf{72.0\%} \\
      \bottomrule
    \end{tabular}
  }
\end{table}

\section{Discussion}

\begin{table}[t]
\centering
\caption{Attack performance comparison of \ours variants on UCF-101.}
\footnotesize
\resizebox{\linewidth}{!}{
\begin{tabular}{lrrrrrr}
\toprule
\multirow{2}{*}[-0.5ex]{Model} &
\multirow{2}{*}[-0.5ex]{Attack} &
\multicolumn{4}{c}{UCF-101} \\
\cmidrule(lr){3-6}
& & ASR$\uparrow$ & TI$\downarrow$ & SSIM$\uparrow$ & PSNR$\uparrow$ \\
\midrule
\multirow{3}{*}{C3D}
& FeatureFool-random  & 53\%  & 3.2534 & 0.8709 & \textbf{29.3015} \\
& FeatureFool-full  & 65\%  & \textbf{3.0564} & 0.8761 & 28.6497 \\
& \textbf{FeatureFool} & \textbf{70\%} & 3.1664 & \textbf{0.8834} & 29.0297 \\
\midrule
\multirow{3}{*}{I3D}
& FeatureFool-random  & 57\%  & 3.7934 & 0.8903 & 29.3648 \\
& FeatureFool-full  & 70\%  & 3.5652 & 0.8864 & 30.0031 \\
& \textbf{FeatureFool} & \textbf{74\%} & \textbf{3.2937} & \textbf{0.8914} & \textbf{30.0137} \\
\bottomrule
\end{tabular}
}
\label{tab:attack_performance_variants_ucf101}
\end{table}
\subsection{Why Use Maximum Optical Flow?}
\begin{figure}
\centering
	\captionsetup{
		font={scriptsize}, 
	}
	\begin{adjustbox}{valign=t}
		\includegraphics[width=0.95\linewidth]{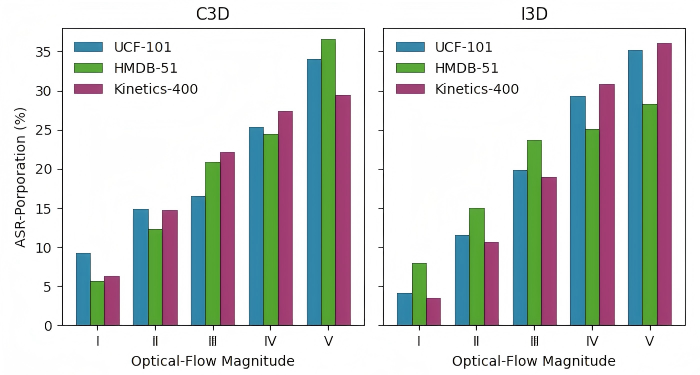}
	\end{adjustbox}
	\caption{ASR proportion versus optical-flow level (I–V); higher flow yields more successful attack frames.}
	\label{figs:optical_flow}
    \vspace{-3mm}
\end{figure}

\begin{figure}
\centering
	\captionsetup{
		font={scriptsize}, 
	}
	\begin{adjustbox}{valign=t}
		\includegraphics[width=0.95\linewidth]{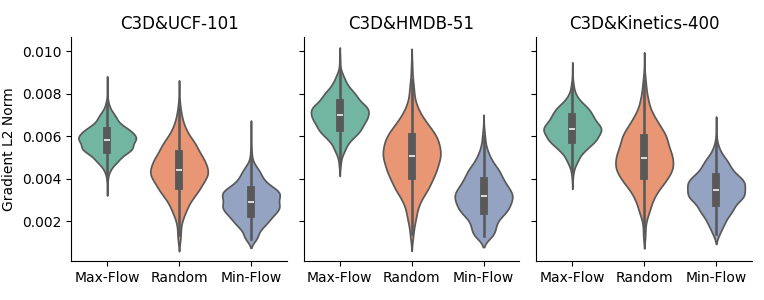}
	\end{adjustbox}
	\caption{Normalised GB-gradient $L_{2}$-norm distributions across frames for three C3D-trained datasets. The distributions of Max-Flow frames are consistently shifted toward higher gradient magnitudes, validating their use as a proxy for the most model-sensitive locations in a black-box setting.}
	\label{figs:optical_grad_info_c3d}
    \vspace{-3mm}
\end{figure}

To explore the impact of optical-flow values on attacks, we select 50 successful attack-videos from each dataset and use every frame to attack each classifier. In Figure~\ref{figs:optical_flow}, x-axis I–V represent uniformly increasing optical-flow levels. The vertical axis shows the proportion of successful-attack frames originating from each optical-flow level for the same video, averaged over the 50 videos. Across the three datasets and two classifiers, we observe that the proportion of successful-attack frames increases with the optical-flow magnitude. This indicates that, for a given video, selecting the frame with higher optical flow yields a higher ASR than choosing frames with lower flow values. The gradient magnitudes further corroborate this observation. Figure~\ref{figs:optical_grad_info_c3d} shows that the distribution of gradient norms for Max-Flow frames is consistently shifted higher than that of other flow frames, indicating richer discriminative information. A larger GB-gradient in the attack-video implies that we extract a more salient pattern from the source DNN. Broadcasting this pattern as a universal perturbation template over all frames of the victim video enables efficient fooling of the black-box model without accessing its gradients (cross-evaluation).

\begin{figure}
\centering
	\captionsetup{
		font={scriptsize}, 
	}
	\begin{adjustbox}{valign=t}
		\includegraphics[width=0.7\linewidth]{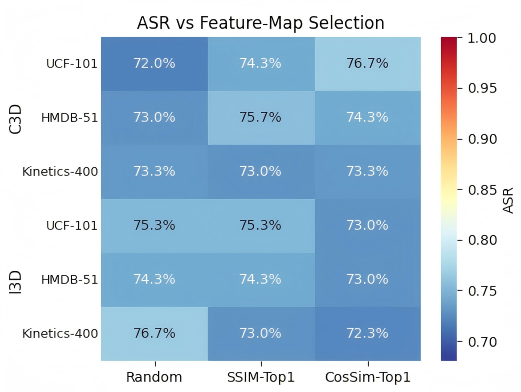}
	\end{adjustbox}
	\caption{Performance of \ours under different attack-video selection strategies. Repeated selection three times per strategy to compute ASR.
}
	\label{figs:selection_heatmap}
    \vspace{-3mm}
\end{figure}

\subsection{Possible Variants of FeatureFool}
\textbf{Attack-Video Selection.}
We further examine whether \ours benefits from carefully selecting the attack-video. As shown in Figure~\ref{figs:selection_heatmap}, randomly picking the attack-video yields the same ASR ($\approx$ 70\%) as choosing the most SSIM- or cosine-similar clip to the target. Thus, no prior selection is needed, a random source video is sufficient for cross-domain attacks.

\textbf{Frame Selection.} Taking UCF-101 as an example, we explore two variants: computing the feature map on a randomly selected frame (\ours-random) and computing it on every frame (\ours-full). In Table~\ref{tab:attack_performance_variants_ucf101}, the ASR results show that \ours-random performs significantly worse than both \ours-full and the original \ours, while \ours-full is close but still inferior to \ours. This demonstrates that combining maximum-optical-flow frame selection with GB to extract the key feature map has a positive and non-trivial effect. Moreover, \ours is regarded as a simpler and more efficient method, as it eliminates the need to compute a feature map for every frame when dealing with long videos. Please refer to the Sup. Mat \ref{sec:Experiments}. for more variants performance.

\begin{figure}[ht]
\centering
	\captionsetup{
			font={scriptsize}, 
		}
	\begin{adjustbox}{valign=t}
	\includegraphics[width=0.95\linewidth]{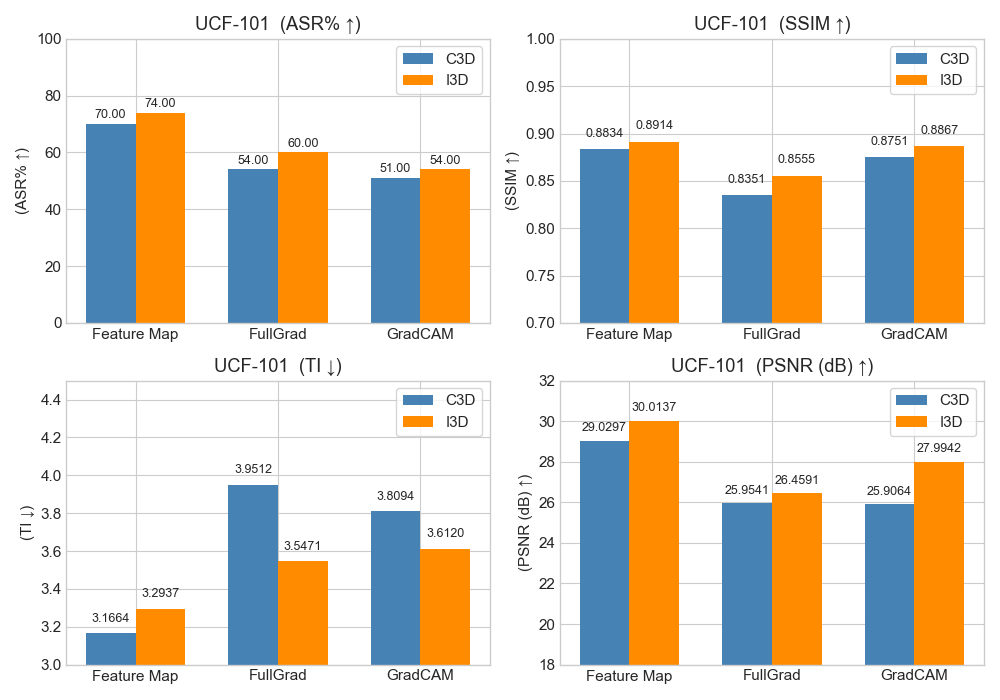}
	\end{adjustbox}
	\caption{The performance of different noise types on UCF-101.}
	\label{figs:noise_ucf101}
    \vspace{-3mm}
\end{figure}

\textbf{Perturbation Types Selection.}
Regarding the choice of perturbation type, we compared FullGrad \cite{srinivas2019full} and GradCam \cite{selvaraju2016grad} against the raw feature map used. Figure~\ref{figs:noise_ucf101} shows that feature maps with richer semantic representations yield more strong perturbations for clean videos and achieve better video-quality performance, benefiting from the finer-grained information it carries compared with other attention maps. Additional results please refer to the Sup. Mat \ref{sec:Experiments}.

\subsection{Visual Interpretation}
\begin{figure}
\centering
	\captionsetup{
		font={scriptsize}, 
	}
	\begin{adjustbox}{valign=t}
		\includegraphics[width=0.85\linewidth]{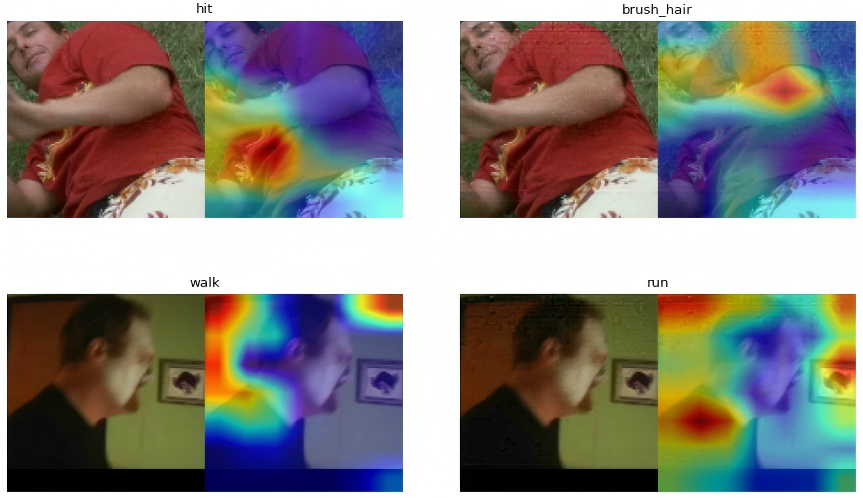}
	\end{adjustbox}
	\caption{Grad-CAM attention visualization of (left) clean video frames and (right) the same frames after \ours attack.}
	\label{figs:attack_visual}
    \vspace{-3mm}
\end{figure}

To visualize how \ours influences model attention, we apply Grad-CAM \cite{selvaraju2016grad} to frames before and after the attack. As shown in Figure~\ref{figs:attack_visual}, the left column displays the attention maps of the clean video, while the right column shows the same frames after the \ours attack. Clearly, the model’s attention distribution changes dramatically; this is largely attributed to the feature map produced by \ours, which embeds DNN-extracted information. Such information directly alters the model’s attention, leading to successful fooling.

\subsection{Impact of Feature-Map Injection}
\begin{figure}
\centering
	\captionsetup{
		font={scriptsize}, 
	}
	\begin{adjustbox}{valign=t}
		\includegraphics[width=0.95\linewidth]{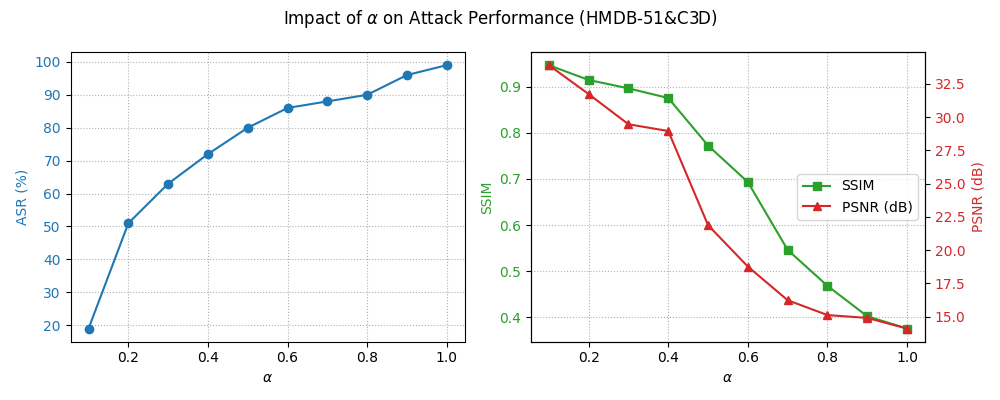}
	\end{adjustbox}
	\caption{Impact of different $\alpha$ intensities on ASR, SSIM and PSNR (HMDB-51$\&$C3D).}
	\label{figs:alpha_HMDB-51&C3D}
    \vspace{-3mm}
\end{figure}

Taking HMDB-51 with the C3D classifier as an example, we varied $\alpha$ from 0.1 to 1.0 in steps of 0.1 and recorded the resulting ASR, SSIM, and PSNR. Intuitively, as the injection strength $\alpha$ increases, the feature map produced by Guided Back-propagation exerts a stronger influence on the clean video. As shown in Figure~\ref{figs:alpha_HMDB-51&C3D}: ASR rises monotonically, while SSIM and PSNR drop monotonically. To balance attack performance and imperceptibility, an appropriate $\alpha$ is required; we set $\alpha$  = 0.4 in our experiments. A visual comparison is provided in
the Sup. Mat \ref{sec:Experiments}.

\subsection{Defense mechanisms}
\begin{table}[t]
\centering
\caption{Residual-ASR ($\uparrow$) performance against defense.}
\footnotesize
\resizebox{0.95\linewidth}{!}{
\begin{tabular}{cccccccc}
\toprule
\multirow{2}{*}[-0.5ex]{Model} & 
\multirow{2}{*}[-0.5ex]{Attack} & \multicolumn{2}{c}{UCF-101} & \multicolumn{2}{c}{HMDB-51} & \multicolumn{2}{c}{Kinetics-400} \\
\cmidrule(r){3-4}\cmidrule(r){5-6}\cmidrule(r){7-8}
 & & TS & DPs & TS & DPs & TS & DPs\\
\midrule
\multirow{5}{*}{C3D} 
& Sparse-RS \cite{croce2022sparse} & 42.0\%  & 39.0\% & 33.0\% & 34.0\%& 41.0\% &  44.0\%\\
& Adv-watermark \cite{jia2020adv} & 44.0\% & 42.0\% & 36.0\% & 42.0\%  & 41.0\% & 48.0\%\\
& PatchAttack \cite{yang2020patchattack} & 55.0\% & 46.0\% & 43.0\% & 51.0\%  & 49.0\% & 56.0\%\\
& BSC \cite{chen2022attacking} & 51.0\% & 46.0\% & 50.0\% & 49.0\%  & 48.0\% & 52.0\%\\
& \textbf{FeatureFool}  & \textbf{66.0\%} & \textbf{64.0\%} & \textbf{65.0}\%  & \textbf{67.0\%} & \textbf{62.0\%} & \textbf{64.0\%}\\
\midrule
\multirow{5}{*}{I3D} 
& Sparse-RS \cite{croce2022sparse}  & 42.0\% & 44.0\% & 38.0\% & 42.0\% & 47.0\% & 43.0\%\\
& Adv-watermark \cite{jia2020adv}  & 41.0\% & 46.0\% & 40.0\% & 45.0\% &  48.0\% & 49.0\%\\
& PatchAttack \cite{yang2020patchattack} & 49.0\% & 53.0\% & 51.0\% & 48.0\%  & 55.0\% & 53.0\%\\
& BSC \cite{chen2022attacking} & 46.0\% & 49.0\% & 52.0\% & 49.0\%  & 51.0\% & 48.0\%\\
& \textbf{FeatureFool}  & \textbf{65.0\%} & \textbf{66.0\%} & \textbf{64.0\%} & \textbf{68.0\%} & \textbf{63.0\%} & \textbf{67.0\%}\\
\bottomrule
\end{tabular}
}
\label{tab:defense_performance}
\vspace{-3mm}
\end{table}
We reproduced two video-specific defenses: (i) \textbf{Defense Patterns (DPs)} \cite{lee2023defending}, which overlay learned patterns to push adversarial videos back to the correct class, 
\begin{equation}
\mathbf{x}_{\text{defend}} = \mathbf{x} + d,\quad d\leftarrow\text{generated from a pre-trained model},
\end{equation}
and (ii) \textbf{Temporal Shuffling (TS)} \cite{hwang2024temporal}, which reorder frames to destroy adversarial perturbations while not being critical to the clean original video.
\begin{equation}
\mathbf{x}_{\text{defend}} = \text{ShuffleFrames}(\mathbf{x},\,h_{1},\,h_{2}).
\end{equation}
Here, $h_1$ and $h_2$ are the hyper-parameters selected for the TS defense mechanism. We feed 100 adversarial videos produced by \ours into each defense and measure the residual ASR---the fraction of still-successful attacks after defense.  
\begin{equation}
\text{Residual-ASR} = \mathbb{I}\!\left[f(\mathbf{x}_{\text{defend}})\neq y\right].
\end{equation}
Table~\ref{tab:defense_performance} summarises the results. More than 60\% of the 100 adversarial videos generated by \ours still fool the corresponding classifier after the defense is applied, whereas Sparse-RS and Adv-watermark achieve only around 40\% success under the same two advanced defenses.  

We attribute this reason to the semantic nature of \ours: its gradient map is anchored to the motion-richest frame and broadcast globally, so the adversarial signature is spatially consistent in every frame.  
Consequently, DPs template cannot fully cancel a globally coherent, motion-aligned perturbation. TS exhibits a similar performance: although frame order is corrupted, the same adversarial feature map remains in each frame, so the shuffled clip still lies on the wrong side of the decision boundary. Consequently, these two advanced defenses still struggle to cope with \ours in most scenarios.

\section{Conclusion}
We present \ours, the first \textbf{zero-query}, black-box adversarial attack that uses a feature map to operate directly on the clean-video in video systems. By coupling Maximum-Optical-Flow frame selection with Guided Back-propagation, our method crafts a single, motion-rich feature map that is broadcast to every frame, bypassing the need for iterative optimization or training. Extensive experiments on 
three benign datasets, and harmful-video clips demonstrate that \ours achieves \textgreater{}70 \% ASR against C3D and I3D, while maintaining high visual fidelity (SSIM \textgreater{} 0.87, PSNR \textgreater{} 28 dB). And demonstrates strong robustness against two advanced video-level defenses, namely DPs and TS. Moreover, the transferable perturbation evades the safety filters of VideoLLaMA2 and ShareGPT4Video in \textgreater{}70 \% of cases and even can trigger hallucination. These findings highlight the urgent need for robust defenses for both traditional video classifiers and emerging Video-LLMs.

{
    \small
    \bibliographystyle{ieeenat_fullname}
    \bibliography{main}
}

\clearpage
\setcounter{page}{1}
\maketitlesupplementary

\section{More Details on Victim Models}
\label{sec:Victim Models}
\textbf{Pre-trained Video Classifiers.} To obtain the source-classification models used in our study, we first performed standard supervised pre-training on the training partitions of UCF-101 \cite{soomro2012ucf101} and HMDB-51 \cite{kuehne2011hmdb}. For the substantially larger Kinetics-400 \cite{kay2017kinetics} dataset, we adopted the official checkpoints released by the MXNet \cite{chen2015mxnet} project instead of retraining. Owing to disparate preprocessing conventions across benchmarks, the two backbones operate under different spatio-temporal resolutions: I3D \cite{carreira2017quo} ingests 32-frame clips at 224$\times$224 pixels when evaluated on Kinetics-400, whereas C3D \cite{tran2015learning} and every other dataset/backbone pair processes 16-frame stacks of size 112$\times$112. The resulting recognition accuracies are summarised in Table~\ref{tab:pretrain_acc}.

\begin{table}[t]
\centering
\caption{Top-1 accuracy (\%) of C3D and I3D after pre-training.}
\label{tab:pretrain_acc}
\resizebox{0.5\linewidth}{!}{
\begin{tabular}{lcc}
\toprule
Dataset  & C3D  & I3D \\
\midrule
UCF-101  & 83.54\% & 61.70\% \\
HMDB-51  & 66.77\% & 47.92\% \\
Kinetics-400 & 59.50\% & 71.80\% \\
\bottomrule
\end{tabular}
}
\end{table}

\section{Computing Maximum-Optical-Flow Frame with Farneback Algorithm}
With Farneback Algorithm \cite{farneback2003two}, given an input video $\mathbf{X}\in\mathbb{R}^{T\times H\times W\times 3}$, we first convert it to grayscale and down-sample to $0.5\times$ resolution to reduce computational cost. Let $\mathbf{X}_{t}^{\text{gray}}\in\mathbb{R}^{h\times w}$ denote the $t$-th grayscale frame, where $h=\lfloor H/2\rfloor$ and $w=\lfloor W/2\rfloor$.

\paragraph{Polynomial Expansion.}
For each pixel $\mathbf{p}=(x,y)$ we fit a quadratic polynomial inside a $5\times 5$ neighbourhood $\mathcal{N}(\mathbf{p})$ by weighted least squares:
\begin{equation}
I(\mathbf{q})\approx\mathbf{q}^{\top}\!\mathbf{A}(\mathbf{p})\,\mathbf{q}+\mathbf{b}(\mathbf{p})^{\top}\mathbf{q}+c(\mathbf{p}),
\\ \qquad\mathbf{q}\in\mathcal{N}(\mathbf{p}),
\end{equation}
where $\mathbf{A}(\mathbf{p})\in\mathbb{R}^{2\times2}$ is symmetric matrix, $\mathbf{b}(\mathbf{p})\in\mathbb{R}^{2}$ is linear, and $c(\mathbf{p})$ is constant. Weights are given by a 2-D Gaussian window $w_{\sigma}(\mathbf{q})=\exp\!\bigl(-\|\mathbf{q}-\mathbf{p}\|^{2}/(2\sigma^{2})\bigr)$; OpenCV \cite{bradski2000opencv} uses $\sigma=1.2$ by default.

\paragraph{Two-Frame Displacement Constraint.}
Let the polynomial coefficients of two successive frames be $(\mathbf{A}_{t-1},\mathbf{b}_{t-1})$ and $(\mathbf{A}_{t},\mathbf{b}_{t})$. Under the local translation assumption $\mathbf{A}_{t-1}\approx\mathbf{A}_{t}\triangleq\mathbf{A}$, the displacement vector $\mathbf{d}(\mathbf{p})=[\Delta u,\Delta v]^{\top}$ satisfies
\begin{equation}
\mathbf{A}(\mathbf{p})\,\mathbf{d}(\mathbf{p})=\frac{1}{2}\bigl[\mathbf{b}_{t-1}(\mathbf{p})-\mathbf{b}_{t}(\mathbf{p})\bigr].
\end{equation}
\begin{equation}
\mathbf{d}(\mathbf{p})=
\Big[\!\sum_{\mathbf{q}\in\Omega}\!w\mathbf{A}^{\top}\mathbf{A}\Big]^{-1}
\sum_{\mathbf{q}\in\Omega}\!w\mathbf{A}^{\top}
\frac{\mathbf{b}_{t-1}-\mathbf{b}_{t}}{2}.
\end{equation}

\paragraph{Flow Magnitude.}
For the frame pair $(t-1,t)$ we obtain the dense flow field $\mathcal{F}_{t}(\mathbf{p})=\mathbf{d}(\mathbf{p})$ and compute its average magnitude:
\begin{equation}
m_{t}=\frac{1}{hw}\sum_{\mathbf{p}\in\Omega}\|\mathcal{F}_{t}(\mathbf{p})\|_{2}=\frac{1}{hw}\sum_{x,y}\sqrt{(\Delta u)^{2}+(\Delta v)^{2}}.
\end{equation}
Boundary handling: $m_{0}=m_{1}$ and $m_{T}=m_{T-1}$.

\paragraph{Maximum-Optical-Flow Frame Index.}
Finally we select
\begin{equation}
t^{*}=\arg\max_{t=0,\dots,T}\;m_{t}.
\end{equation}
This frame is used by \ours for Guided Back-propagation.

\section{More Experiments Results}
\label{sec:Experiments}
\subsection{Distributions of Flow frames.}
\begin{figure}
\centering
	\captionsetup{
		font={scriptsize}, 
	}
	\begin{adjustbox}{valign=t}
		\includegraphics[width=0.95\linewidth]{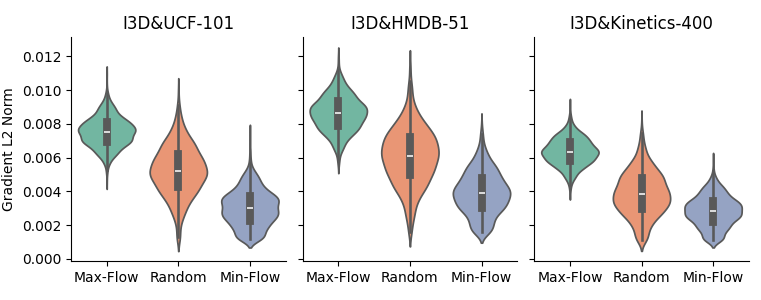}
	\end{adjustbox}
	\caption{Normalised GB-gradient $L_{2}$-norm distributions across frames for three C3D-trained datasets. The distributions of Max-Flow frames are consistently shifted toward higher gradient magnitudes, validating their use as a proxy for the most model-sensitive locations in a black-box setting.}
	\label{figs:optical_grad_info_i3d}
    \vspace{-3mm}
\end{figure}

As evidenced on the I3D \cite{carreira2017quo}-family models in Figure~\ref{figs:optical_grad_info_i3d}, the gradient-norm distribution of Max-Flow frames is markedly shifted above those of Random- or Min-Flow frames. Therefore, we harvest a stronger decision-sensitive pattern from the I3D source network; broadcasting this pattern as a universal, motion-aligned perturbation across every frame of the victim video effectively misleads the black-box classifier without querying its gradients.

\subsection{Variants Performance.} Table~\ref{tab:attack_performance_variants_hmdb51} and Table~\ref{tab:attack_performance_variants_kinetics400} report the results of variants on HMDB-51 \cite{kuehne2011hmdb} and Kinetics-400 \cite{kay2017kinetics}. Numerical results show that SSIM \cite{wang2004image} and PSNR are still influenced by the frame-selection strategy, yet ASR exhibits a consistent trend, confirming the effectiveness of maximum-optical-flow selection.

\begin{table}[t]
\centering
\caption{Attack performance comparison of FeatureFool variants on HMDB-51.}
\footnotesize
\resizebox{\linewidth}{!}{
\begin{tabular}{lrrrrrr}
\toprule
\multirow{2}{*}[-0.5ex]{Model} &
\multirow{2}{*}[-0.5ex]{Attack} &
\multicolumn{4}{c}{HMDB-51} \\
\cmidrule(lr){3-6}
& & ASR$\uparrow$ & TI$\downarrow$ & SSIM$\uparrow$ & PSNR$\uparrow$ \\
\midrule
\multirow{3}{*}{C3D}
& FeatureFool-random  & 56\%  & 3.4134 & 0.8619 & 29.1335 \\
& FeatureFool-full  & 68\%  & 3.7514  & 0.8637 & \textbf{30.1103} \\
& \textbf{FeatureFool} & \textbf{72\%} & \textbf{3.6821} & \textbf{0.8755} & 28.9602 \\
\midrule
\multirow{3}{*}{I3D}
& FeatureFool-random  & 62\%  & 3.9355 & \textbf{0.8955} & 29.0109 \\
& FeatureFool-full  & 67\%  & 3.9521 & 0.8734 & 28.4061 \\
& \textbf{FeatureFool} & \textbf{73\%} & \textbf{4.3467} & 0.8861 & \textbf{29.4111} \\
\bottomrule
\end{tabular}
}
\label{tab:attack_performance_variants_hmdb51}
\end{table}

\begin{table}[t]
\centering
\caption{Attack performance comparison of FeatureFool variants on Kinetics-400.}
\footnotesize
\resizebox{\linewidth}{!}{
\begin{tabular}{lrrrrrr}
\toprule
\multirow{2}{*}[-0.5ex]{Model} &
\multirow{2}{*}[-0.5ex]{Attack} &
\multicolumn{4}{c}{Kinetics-400} \\
\cmidrule(lr){3-6}
& & ASR$\uparrow$ & TI$\downarrow$ & SSIM$\uparrow$ & PSNR$\uparrow$ \\
\midrule
\multirow{3}{*}{C3D}
& FeatureFool-random  & 55\%  & 3.9100 & 0.8650 & \textbf{29.8505} \\
& FeatureFool-full  & 61\%  & \textbf{3.4200} & 0.8710 & 28.9554 \\
& \textbf{FeatureFool} & \textbf{70\%} & 3.7553 & \textbf{0.8864} & 28.5624 \\
\midrule
\multirow{3}{*}{I3D}
& FeatureFool-random  & 59\%  & 4.1200 & \textbf{0.8840} & 29.6508 \\
& FeatureFool-full  & 66\%  & 3.7800 & 0.8810 & 29.7026 \\
& \textbf{FeatureFool} & \textbf{72\%} & \textbf{3.6594} & 0.8631 & \textbf{30.2497} \\
\bottomrule
\end{tabular}
}
\label{tab:attack_performance_variants_kinetics400}
\end{table}

\begin{figure}[ht]
\centering
	\captionsetup{
			font={scriptsize}, 
		}
	\begin{adjustbox}{valign=t}
	\includegraphics[width=0.95\linewidth]{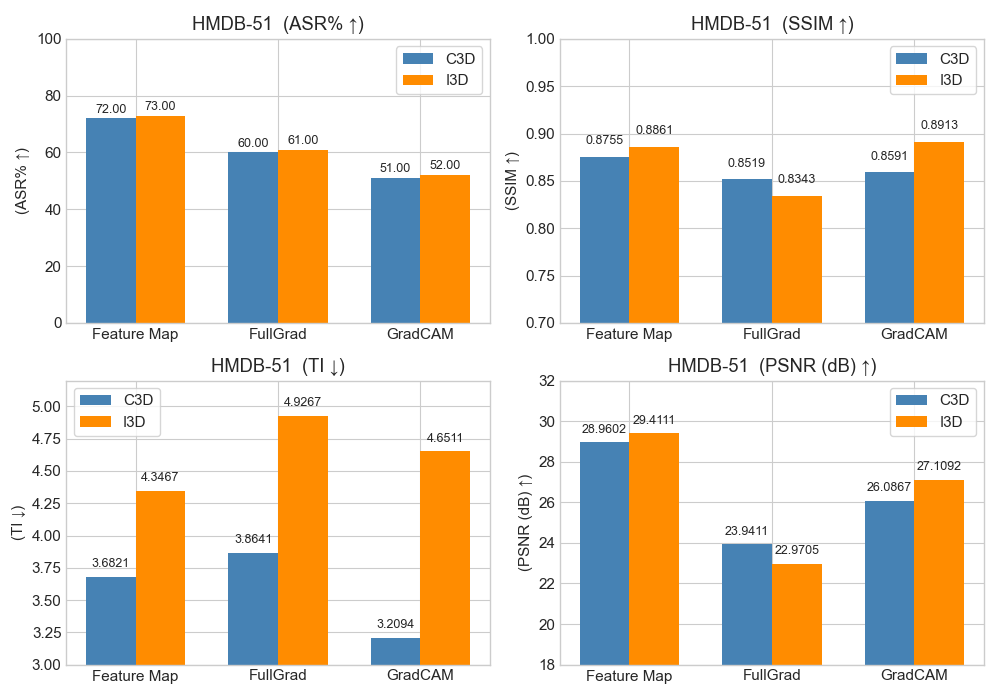}
	\end{adjustbox}
	\caption{The performance of different noise types on HMDB-51.}
	\label{figs:noise_hmdb51}
    \vspace{-3mm}
\end{figure}

\begin{figure}[ht]
\centering
	\captionsetup{
			font={scriptsize}, 
		}
	\begin{adjustbox}{valign=t}
	\includegraphics[width=0.95\linewidth]{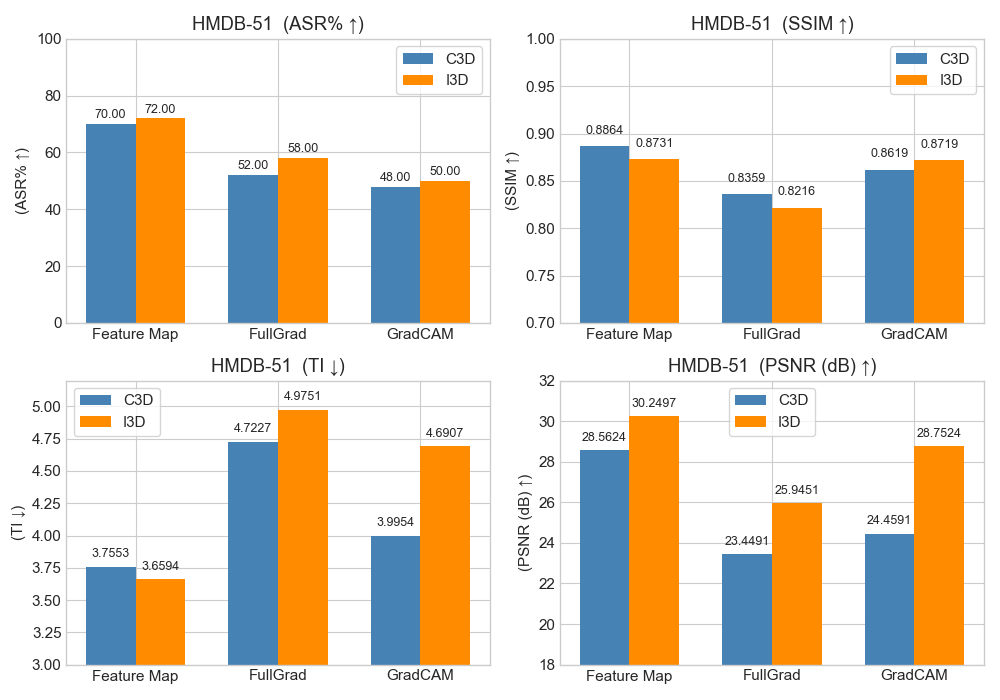}
	\end{adjustbox}
	\caption{The performance of different noise types on Kinetics-400.}
	\label{figs:noise_kinetics400}
    \vspace{-3mm}
\end{figure}

\subsection{Perturbation Selection.}
Figures~\ref{figs:noise_hmdb51} and Figure~\ref{figs:noise_kinetics400} report the behavior of different noise variants on HMDB-51 and Kinetics-400. Again, the feature-map noise achieves the best trade-off between ASR and video quality: it delivers the highest ASR in all cases, benefiting from the finer-grained information it carries compared with other attention maps, and maintains superior adversarial-sample quality in most scenarios.
\begin{figure}[ht]
\centering
	\captionsetup{
			font={scriptsize}, 
		}
	\begin{adjustbox}{valign=t}
	\includegraphics[width=0.95\linewidth]{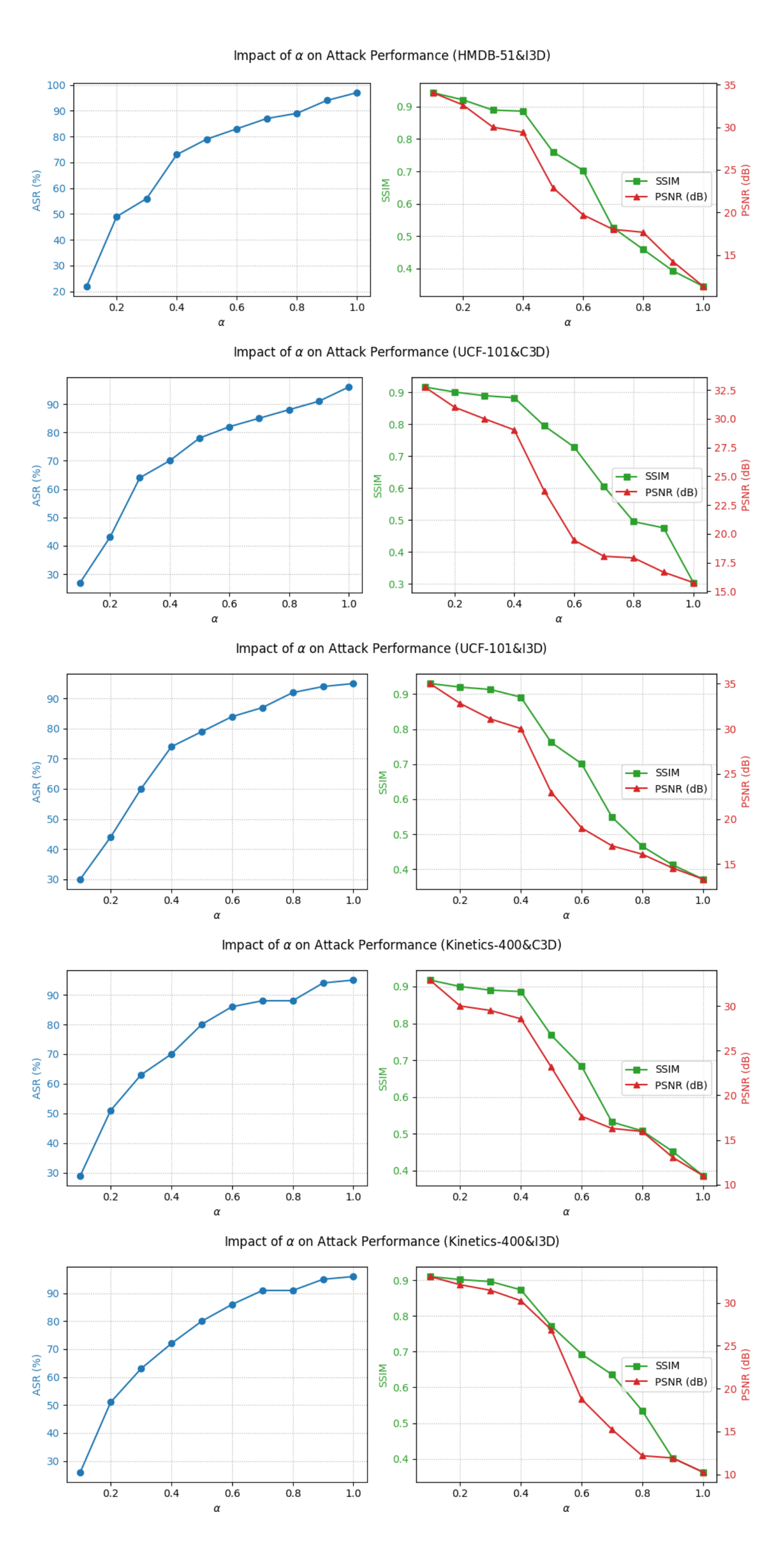}
	\end{adjustbox}
	\caption{More results about impact of different $\alpha$ intensities on ASR, SSIM and PSNR.}
	\label{figs:impact_append}
    \vspace{-3mm}
\end{figure}

\begin{figure}[ht]
\centering
	\captionsetup{
			font={scriptsize}, 
		}
	\begin{adjustbox}{valign=t}
	\includegraphics[width=0.95\linewidth]{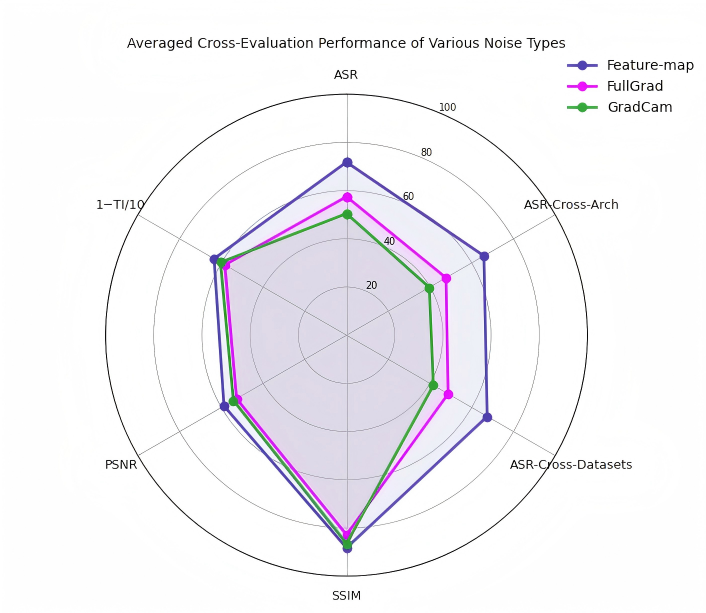}
	\end{adjustbox}
	\caption{Radar-chart comparison of averaged performance across noise types.
ASR: vanilla attack-success rate; ASR-Cross-Arch: cross-model transfer; ASR-Cross-Datasets: cross-dataset transfer; TI inverted for consistency.
Feature-map perturbations (purple) consistently enclose the other polygons, demonstrating superior performance.}
	\label{figs:noise_cross}
    \vspace{-3mm}
\end{figure}

\subsection{Cross-Evaluation across Noise Types.}
We also conducted statistics on the cross-model-architecture and cross-dataset performance of FullGrad \cite{srinivas2019full} and GradCam \cite{selvaraju2016grad}, and averaged the cross-performance. As shown in Figure~\ref{figs:noise_cross}, perturbations generated with the feature map achieve superior attack performance, especially on all ASR-related metrics. This benefit largely stems from the fact that feature maps carry finer-grained, semantically meaningful representations than the other two maps, enabling stronger yet equally imperceptible perturbations.

\subsection{Trade-off between Stealthiness and ASR.} 
\begin{figure}
\centering
	\captionsetup{
		font={scriptsize}, 
	}
	\begin{adjustbox}{valign=t}
		\includegraphics[width=0.95\linewidth]{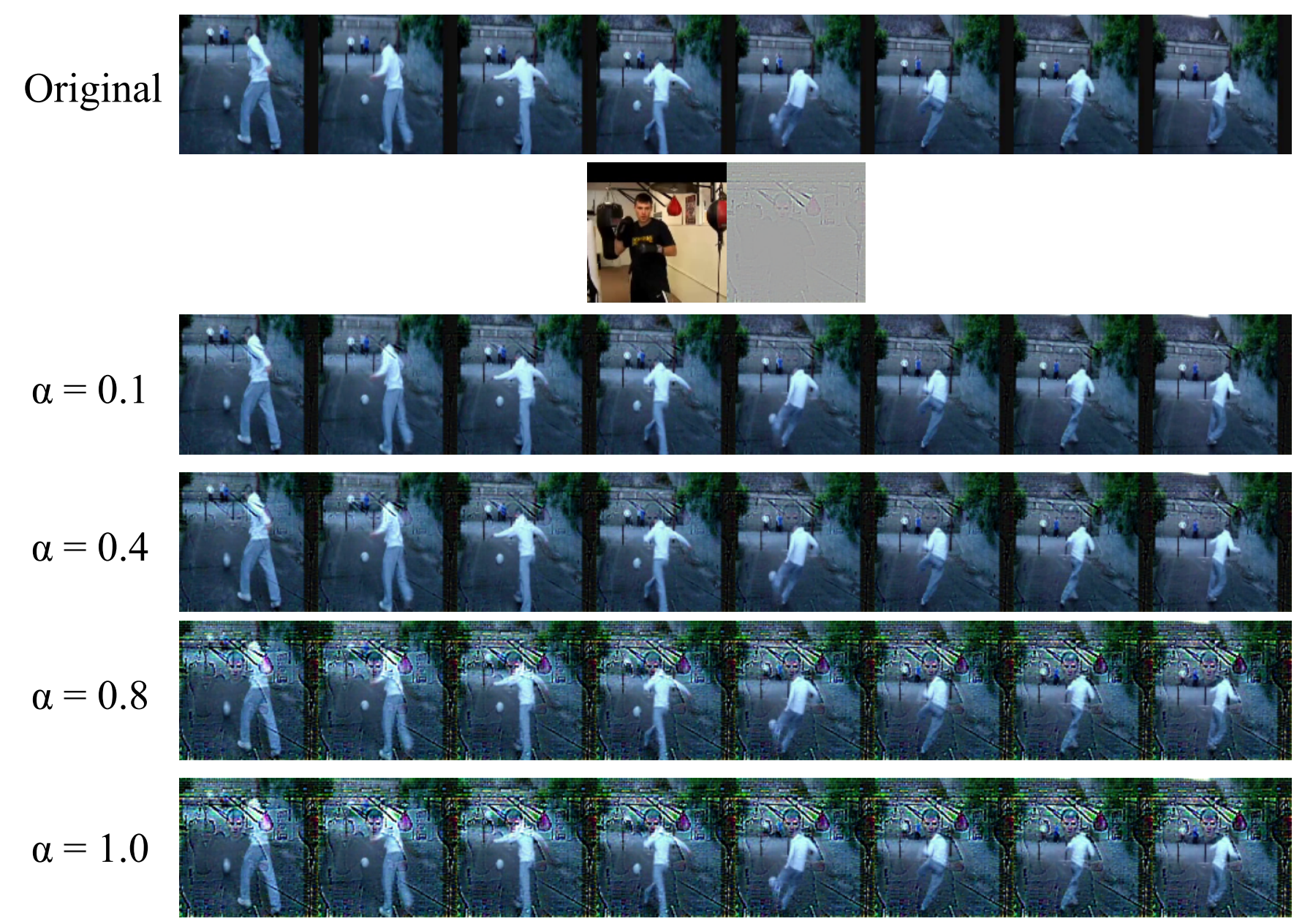}
	\end{adjustbox}
	\caption{Impact of different $\alpha$ intensities on video appearance.}
	\label{figs:alpha_compare}
    \vspace{-3mm}
\end{figure}

We explore the performance of \ours at $\alpha\in\{0.1,0.4,0.8,1.0\}$.
Tables~\ref{tab:attack_performance_alpha_UCF101}, Tables~\ref{tab:attack_performance_alpha_hmdb51} and Tables~\ref{tab:attack_performance_alpha_Kinetics400} present additional results across multiple datasets and models under these settings.
Consistently across all splits, increasing $\alpha$ intensifies the feature-map injection, thereby exerting a stronger influence on the clean video and misleading the classifier.
This gain in attack strength, however, comes at the cost of visual quality; higher $\alpha$ values visibly degrade SSIM and PSNR.
Consequently, one must trade off stealthiness against ASR when setting the attack magnitude. Figures~\ref{figs:impact_append} illustrates the impact of $\alpha$ on UCF-101, HMDB-51, and Kinetics-400 with the corresponding C3D and I3D models. Figure~\ref{figs:alpha_compare} shows the visual comparison.

\begin{table}[t]
\centering
\caption{Attack performance with different $\alpha$ on UCF-101}
\footnotesize
\resizebox{\linewidth}{!}{
\begin{tabular}{lrrrrrr}
\toprule
\multirow{2}{*}[-0.5ex]{Model} &
\multirow{2}{*}[-0.5ex]{Attack} &
\multicolumn{4}{c}{UCF-101} \\
\cmidrule(lr){3-6}
& & ASR$\uparrow$ & TI$\downarrow$ & SSIM$\uparrow$ & PSNR$\uparrow$ \\
\midrule
\multirow{4}{*}{C3D}
& FeatureFool ($\alpha$=0.1)  & 27\%  & 2.4244 & 0.9169 & 32.7569 \\
& FeatureFool ($\alpha$=0.4)  & 70\%  & 3.1664 & 0.8834 & 29.0297 \\
& FeatureFool ($\alpha$=0.8)  & 88\% & 6.7056 & 0.4953 & 17.9106 \\
& FeatureFool ($\alpha$=1.0)  & 96\% & 10.0351 & 0.3044 & 15.7651 \\
\midrule
\multirow{4}{*}{I3D}
& FeatureFool ($\alpha$=0.1)  & 30\%  & 2.1954 & 0.9303 & 35.0024 \\
& FeatureFool ($\alpha$=0.4)  & 74\%  & 3.2937 & 0.8914 & 30.0137 \\
& FeatureFool ($\alpha$=0.8)  & 91\%  & 5.9961 & 0.4657 & 16.1037 \\
& FeatureFool ($\alpha$=1.0)  & 95\%  & 8.9421 & 0.3720 & 13.3473 \\
\bottomrule
\end{tabular}
}
\label{tab:attack_performance_alpha_UCF101}
\end{table}

\begin{table}[t]
\centering
\caption{Attack performance with different $\alpha$ on HMDB-51}
\footnotesize
\resizebox{\linewidth}{!}{
\begin{tabular}{lrrrrrr}
\toprule
\multirow{2}{*}[-0.5ex]{Model} &
\multirow{2}{*}[-0.5ex]{Attack} &
\multicolumn{4}{c}{HMDB-51} \\
\cmidrule(lr){3-6}
& & ASR$\uparrow$ & TI$\downarrow$ & SSIM$\uparrow$ & PSNR$\uparrow$ \\
\midrule
\multirow{4}{*}{C3D}
& FeatureFool ($\alpha$=0.1)  & 19\%  & 3.0244 & 0.9467 & 33.9161 \\
& FeatureFool ($\alpha$=0.4)  & 72\%  & 3.6821 & 0.8755 & 28.9602 \\
& FeatureFool ($\alpha$=0.8)  & 90\% & 6.4592 & 0.4682 & 15.1267 \\
& FeatureFool ($\alpha$=1.0)  & 96\% & 9.6651 & 0.3756 & 14.1134 \\
\midrule
\multirow{4}{*}{I3D}
& FeatureFool ($\alpha$=0.1)  & 22\%  & 2.9181 & 0.9431 & 34.0624 \\
& FeatureFool ($\alpha$=0.4)  & 73\%  & 4.3467 & 0.8861 & 29.4111 \\
& FeatureFool ($\alpha$=0.8)  & 89\%  & 6.6304 & 0.4592 & 17.6642 \\
& FeatureFool ($\alpha$=1.0)  & 97\%  & 8.2304 & 0.3450 & 11.3207 \\
\bottomrule
\end{tabular}
}
\label{tab:attack_performance_alpha_hmdb51}
\end{table}

\begin{table}[t]
\centering
\caption{Attack performance with different $\alpha$ on Kinetics-400}
\footnotesize
\resizebox{\linewidth}{!}{
\begin{tabular}{lrrrrrr}
\toprule
\multirow{2}{*}[-0.5ex]{Model} &
\multirow{2}{*}[-0.5ex]{Attack} &
\multicolumn{4}{c}{Kinetics-400} \\
\cmidrule(lr){3-6}
& & ASR$\uparrow$ & TI$\downarrow$ & SSIM$\uparrow$ & PSNR$\uparrow$ \\
\midrule
\multirow{4}{*}{C3D}
& FeatureFool ($\alpha$=0.1)  & 29\%  & 2.0244 & 0.9177 & 32.9119 \\
& FeatureFool ($\alpha$=0.4)  & 70\%  & 3.7553 & 0.8864 & 28.5624 \\
& FeatureFool ($\alpha$=0.8)  & 88\% & 6.0482 & 0.5079 & 16.1109 \\
& FeatureFool ($\alpha$=1.0)  & 95\% & 10.6271 & 0.3864 & 11.0034 \\
\midrule
\multirow{4}{*}{I3D}
& FeatureFool ($\alpha$=0.1)  & 26\%  & 3.2021 & 0.9105 & 33.0755 \\
& FeatureFool ($\alpha$=0.4)  & 72\%  & 3.6594 & 0.8731 & 30.2497 \\
& FeatureFool ($\alpha$=0.8)  & 91\%  & 6.4201 & 0.5346 & 12.1782 \\
& FeatureFool ($\alpha$=1.0)  & 96\%  & 7.6891 & 0.3628 & 10.2679 \\
\bottomrule
\end{tabular}
}
\label{tab:attack_performance_alpha_Kinetics400}
\end{table}

\begin{figure}
\centering
	\captionsetup{
		font={scriptsize}, 
	}
	\begin{adjustbox}{valign=t}
		\includegraphics[width=1.0\linewidth]{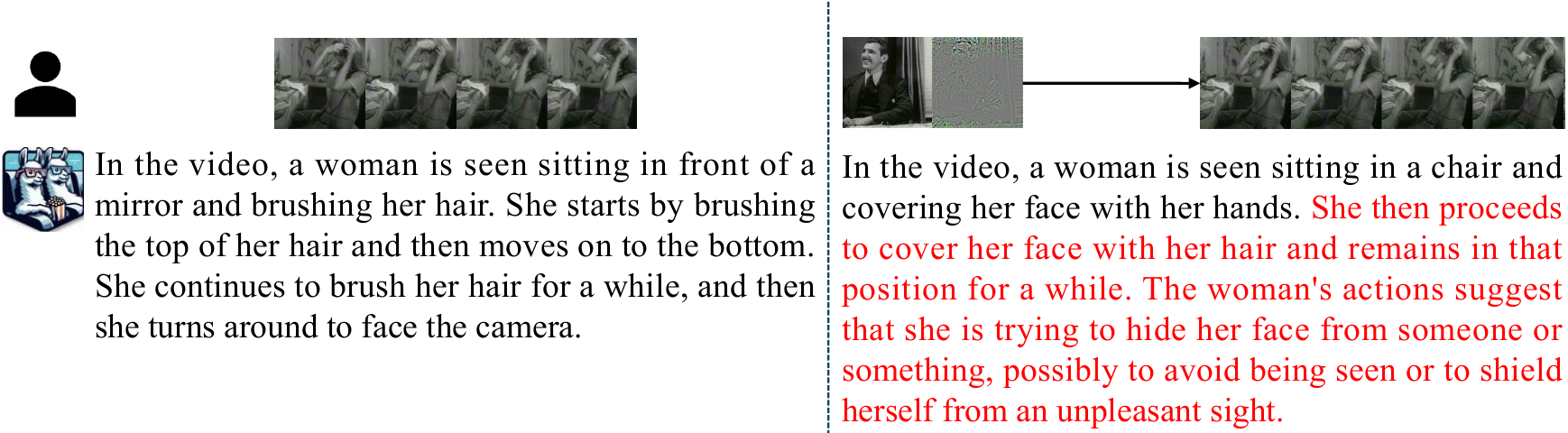}
	\end{adjustbox}
	\caption{Hallucinations induced after \ours perturbation.}
	\label{figs:Hallucination}
    \vspace{-3mm}
\end{figure}

\subsection{Hallucination Showcase.}
\label{sec:Hallucination}
We further observe that \ours can trigger hallucinations in Video-LLMs. An example is given in Figure~\ref{figs:Hallucination}: the injection of external features causes the model to output numerous irrelevant sentences, highlighted in red. One possible explanation is that the stealthy yet powerful feature perturbations introduced by \ours shift the clean video’s feature space, forcing the Video-LLM to interpret the video within an adversarial space and ultimately producing hallucinations.

\section{Details about Metrics.} 
\label{sec:Metrics}
\textbf{Temporal Inconsistency (TI).}
Ruder et al.\ \cite{ruder2018artistic} minimise the squared warping residual
\begin{equation}
\mathcal{L}_{\text{temp}}=\sum c_{k}(x_{k}-\omega_{k})^{2}
\end{equation}
to stabilise stylised videos.  We reuse their pipeline but measure the residual on adversarial videos.  
First, define the occlusion-weighted $\ell_{1}$ error between any two frames
\begin{equation}
\mathcal{E}(x_{t},x_{m})=\frac{1}{HWC}\sum_{c=1}^{C}O_{t,m}^{(c)}\odot\bigl|x_{t}^{(c)}-\mathcal{W}(x_{m}^{(c)})\bigr|,
\end{equation}
where $\mathcal{W}$ is the DeepFlow \cite{weinzaepfel2013deepflow} backward warp and $O_{t,m}$ the forward–backward consistency mask.  
Averaging over the whole clip yields the Temporal-Inconsistency index
\begin{equation}
\text{TI}=\frac{1}{2(T-1)}\sum_{t=2}^{T}\Bigl[\mathcal{E}(x_{t},x_{1})+\mathcal{E}(x_{t},x_{t-1})\Bigr].
\end{equation}
Lower TI $\Rightarrow$ smoother motion; higher TI $\Rightarrow$ adversarial flicker.

\textbf{Structural Similarity (SSIM)}. SSIM \cite{wang2004image} assesses perceptual fidelity by comparing local luminance, contrast and structure between the clean video $\mathbf{X}$ and the adversarial video $\mathbf{X}_{\mathrm{adv}}$ frame-wise, then averaging over time:
\begin{equation}
\mathrm{SSIM}(\mathbf{X},\mathbf{X}_{\mathrm{adv}})=\frac{1}{T}\sum_{t=1}^{T}\mathrm{SSIM}(\mathbf{X}_{t},\mathbf{X}_{\mathrm{adv},t})\in[-1,1],
\end{equation}
where $1$ indicates perfect visual match.

\textbf{Peak Signal-to-Noise Ratio (PSNR)}. PSNR is computed on the $\ell_{2}$ error of the 8-bit pixel space:
\begin{equation}
\mathrm{PSNR}(\mathbf{X},\mathbf{X}_{\mathrm{adv}})=10\log_{10}\frac{255^{2}}{\mathrm{MSE}(\mathbf{X},\mathbf{X}_{\mathrm{adv}})}\;[\mathrm{dB}],
\end{equation}
with $\mathrm{MSE}=\frac{1}{CHWT}\|\mathbf{X}-\mathbf{X}_{\mathrm{adv}}\|_{2}^{2}$.  
Higher PSNR (lower MSE) implies smaller perturbation energy.

\section{Introduction to Video Datasets}
\label{sec:Datasets}
\textbf{UCF-101} \cite{soomro2012ucf101} comprises 13,320 realistic videos distributed across 101 sport and daily-life categories.  The collection is recorded at 25 fps with a spatial resolution of 320 × 240; most clips are 5–10 s long and depict nearly static scenes with stable camera motion.

\textbf{HMDB-51} \cite{kuehne2011hmdb} provides 6,849 video clips from 51 action classes extracted YouTube, google and public databases.  The dataset emphasises natural human motions (e.g., walk, wave, smile) under severe illumination changes, camera jitter and partial occlusions.

\textbf{Kinetics-400} \cite{kay2017kinetics} is a large-scale corpus that contains $\approx$ 240 k training videos and 20 k validation videos spanning 400 human actions.  Clips are sourced from YouTube at 25 fps with an average duration of 10 s; the action taxonomy covers fine-grained motions such as “playing violin” or “mopping floor”.

\textbf{Real-Life Violence Situations Dataset} \cite{elesawy2019real} contains 2,000 YouTube clips, half capturing diverse street-fight scenes and half everyday non-violent actions, that serve as realistic positive and negative samples for violence detection.

\textbf{UCF-Crime Dataset} \cite{sultani2018real} is the first large-scale dataset for real-world anomaly detection, offering 128 hours of untrimmed surveillance video that covers 13 realistic anomalies such as abuse, fighting, robbery and vandalism.

\section{Important symbols}
Table~\ref{tab:notation} lists the symbols frequently used in the main paper for quick reference.
\begin{table}[h]
\centering
\caption{Frequently-used symbols in the main paper.}
\label{tab:notation}
\small
\begin{tabular}{@{}cl@{}}
\toprule
Symbol & Description \\ \midrule
$\mathbf{X}\in\mathbb{R}^{T\times C\times H\times W}$ & clean video clip \\
$\mathbf{X}_{\mathrm{att}}$ & video that produces feature map (perturbation) \\
$\mathbf{X}_{\mathrm{adv}}$ & adversarial video \\
$\delta$ & universal perturbation ($\|\delta\|_{\infty}\le\varepsilon$) \\
$t^{*}$ & index of the Max-Optical-Flow frame \\
$\mathbf{G}\in\mathbb{R}^{H\times W\times C}$ & Guided-Backprop feature map \\
$\alpha$ & injection strength of $\mathbf{G}$ \\
$\phi(\cdot;\theta)$ & 3D-CNN classifier (C3D / I3D) \\
$\mathrm{ASR}$ & attack success rate (\%) \\
$\mathrm{TI}\downarrow$ & temporal-inconsistency \\ 
$\mathrm{(S)}$ & the source model in cross-evaluation \\
$\mathrm{(V)}$ & the victim model in cross-evaluation \\ \bottomrule
\end{tabular}
\end{table}

\section{Ethics Statement}
Experiments on violence \cite{elesawy2019real}, crime \cite{sultani2018real} and pornography clips are conducted solely to evaluate model safety. Adversarial perturbations do not create or intensify harmful content; Source videos are public, de-identified, audio-removed, and never re-distributed in adversarial form. Only aggregate metrics are reported. We encourage follow-up work on countermeasures and explicitly discourage any malicious reuse.

\section{Future Work}
\ours demonstrates effective and strong performance only zero-query in untargeted attacks within the video domain. Looking forward, future attention will shift to leveraging feature-map priors for targeted video attacks under zero- or few-query budgets. Moreover, we also care about the hallucinations (Sec.~\ref{sec:Hallucination}) in Video-LLMs that caused by feature-map injection, and investigate whether the hallucinations arise from attention sink \cite{wang2025mirage} induced by the feature maps produced by \ours. Most importantly, future work will delve into more effective defenses against this class of feature-based, stealthy perturbations.


\end{document}